\documentclass[sigconf]{acmart}
\AtBeginDocument{%
  \providecommand\BibTeX{{%
    \normalfont B\kern-0.5em{\scshape i\kern-0.25em b}\kern-0.8em\TeX}}}

\usepackage{float}
\usepackage{newfloat}
\usepackage{booktabs}
\usepackage{multirow}
\usepackage{makecell}
\usepackage{xcolor} % colors
\usepackage{colortbl}
\usepackage{dsfont}
\usepackage{amsmath}
\usepackage{enumitem}
\usepackage{fontawesome}

\usepackage[normalem]{ulem}
\useunder{\uline}{\ul}{}

\definecolor{myyellow}{rgb}{1,1, 0.6}
\definecolor{myorange}{rgb}{1, 0.8, 0.6}
\definecolor{myred}{rgb}{1, 0.6, 0.6}
\definecolor{deepred}{RGB}{200,0,0} % Deep red color
\definecolor{second}{HTML}{FFDAB9}
\definecolor{best}{HTML}{FFC1C1}

\usepackage{multirow}
\usepackage{colortbl}
\usepackage[normalem]{ulem}
\usepackage{bbding}
\useunder{\uline}{\ul}{}
\usepackage{stackrel}
\usepackage{todonotes}

\usepackage{booktabs}
\usepackage{multirow}
\usepackage{makecell}
\usepackage{xcolor}
\usepackage{colortbl}
\usepackage{dsfont}
\usepackage{amsmath}
\usepackage{enumitem}
\usepackage{fontawesome}
\usepackage[ruled,vlined]{algorithm2e}
\usepackage{stfloats}

\usepackage[skins,breakable]{tcolorbox}
\tcbuselibrary{listings}

\usepackage{listings}
\usepackage{tcolorbox}
\usepackage{mdframed}
\usepackage{balance}

\makeatletter
\def\UrlAlphabet{%
      \do\a\do\b\do\c\do\d\do\e\do\f\do\g\do\h\do\i\do\j%
      \do\k\do\l\do\m\do\n\do\o\do\p\do\q\do\r\do\s\do\t%
      \do\u\do\v\do\w\do\x\do\y\do\z\do\A\do\B\do\C\do\D%
      \do\E\do\F\do\G\do\H\do\I\do\J\do\K\do\L\do\M\do\N%
      \do\O\do\P\do\Q\do\R\do\S\do\T\do\U\do\V\do\W\do\X%
      \do\Y\do\Z}
\def\UrlDigits{\do\1\do\2\do\3\do\4\do\5\do\6\do\7\do\8\do\9\do\0}
\g@addto@macro{\UrlBreaks}{\UrlOrds}
\g@addto@macro{\UrlBreaks}{\UrlAlphabet}
\g@addto@macro{\UrlBreaks}{\UrlDigits}
\makeatother

\copyrightyear{2025}
\acmYear{2025}
\setcopyright{acmlicensed}
\acmConference[MM '25]{Proceedings of the 33rd ACM International Conference on Multimedia}{October 27--31, 2025}{Dublin, Ireland}
\acmBooktitle{Proceedings of the 33rd ACM International Conference on Multimedia (MM '25), October 27--31, 2025, Dublin, Ireland}
\acmDOI{10.1145/3746027.3754766}
\acmISBN{979-8-4007-2035-2/2025/10}

\acmSubmissionID{2025-520}

\begin{document}

\title{Robust Modality-Incomplete Anomaly Detection:\\a Modality-Instructive Framework with Benchmark}
% \author{Anonymous Authors}
\author{Bingchen Miao}
\email{miaobingchen23@zju.edu.cn}
\affiliation{%
  \institution{Zhejiang University}
  \city{Hangzhou}
  \state{Zhejiang}
  \country{China}
}

\author{Wenqiao Zhang}
\authornote{Corresponding Author.}
\email{wenqiaozhang@zju.edu.cn}
\affiliation{%
  \institution{Zhejiang University}
  \city{Hangzhou}
  \state{Zhejiang}
  \country{China}
}

\author{Juncheng Li}
\authornotemark[1]
\email{junchengli@zju.edu.cn}
\affiliation{%
  \institution{Zhejiang University}
  \city{Hangzhou}
  \state{Zhejiang}
  \country{China}
}

\author{Wangyu Wu}
\email{v11dryad@foxmail.com}
\affiliation{%
  \institution{University of Liverpool}
  \city{Liverpool}
  \country{England}
}

\author{Siliang Tang}
\email{siliang@zju.edu.cn}
\affiliation{%
  \institution{Zhejiang University}
  \city{Hangzhou}
  \state{Zhejiang}
  \country{China}
}

\author{Zhaocheng Li}
\email{3200106072@zju.edu.cn}
\affiliation{%
  \institution{Zhejiang University}
  \city{Hangzhou}
  \state{Zhejiang}
  \country{China}
}

\author{Haochen Shi}
\email{haochen.shi@umontreal.ca}
\affiliation{%
  \institution{Université de Montréal}
  \city{Montréal}
  \country{Canada}
}

\author{Jun Xiao}
\email{junx@cs.zju.edu.cn}
\affiliation{%
  \institution{Zhejiang University}
  \city{Hangzhou}
  \state{Zhejiang}
  \country{China}
}

\author{Yueting Zhuang}
\email{yzhuang@zju.edu.cn}
\affiliation{%
  \institution{Zhejiang University}
  \city{Hangzhou}
  \state{Zhejiang}
  \country{China}
}

\renewcommand{\shortauthors}{Bingchen Miao et al.}

\begin{abstract}
\label{sec:abstract}

Multimodal Industrial Anomaly Detection (MIAD)—fusing 3D point clouds and 2D RGB for product defect detection—is critical to quality inspection. However, existing MIAD methods assume all modalities are available and paired, overlooking real-scenario modality-missing and risking overfitting to incomplete data. To address these, we conduct the first comprehensive study on \textbf{\underline{M}odality-\underline{I}ncomplete \underline{I}ndustrial \underline{A}nomaly \underline{D}etection (MIIAD)} and establish \textbf{\textit{MIIAD Bench}}, a benchmark covering diverse missing settings. Meanwhile, we propose \textbf{RADAR}, a robust two-stage \textbf{\underline{R}}obust mod\textbf{\underline{A}}lity-instructive fusing and \textbf{\underline{D}}etecting fr\textbf{\underline{A}}mewo\textbf{\underline{R}}k. \textbf{RADAR} integrates \textbf{i)} a Modality-Incomplete Instruction mechanism—guiding the multimodal Transformer to focus more on available modal info, and \textbf{ii)} a Double-Pseudo Hybrid Module to highlight unique modality combinations and reduce overfitting. Our results show \textbf{RADAR} outperforms prior methods markedly on \textbf{\textit{MIIAD Bench}}.

\end{abstract}
% \begin{abstract}
%   A clear and well-documented \LaTeX\ document is presented as an
%   article formatted for publication by ACM in a conference proceedings
%   or journal publication. Based on the ``acmart'' document class, this
%   article presents and explains many of the common variations, as well
%   as many of the formatting elements an author may use in the
%   preparation of the documentation of their work.
% \end{abstract}




% \keywords{Do, Not, Us, This, Code, Put, the, Correct, Terms, for,


\maketitle

\section{Introduction}
\label{sec:introduction}
\newcommand{\blue}[1]{\textcolor{blue}{#1}}
\vspace{-2pt}

\begin{figure}[t]
\vspace{10pt}
\centering
% \hspace{-0.2cm}
\includegraphics[width=0.44\textwidth]{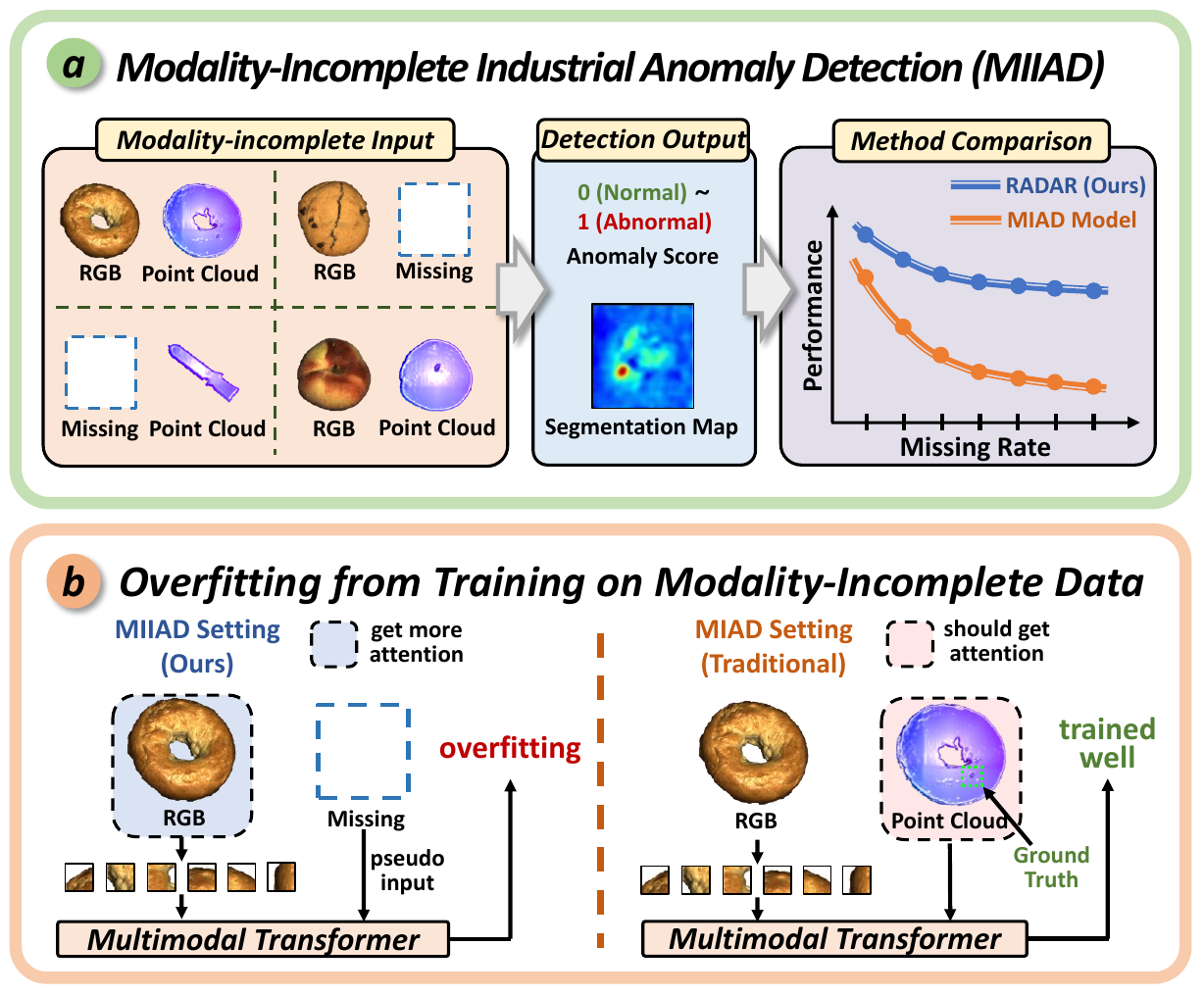}
\vspace{-10pt}
\centering\caption{(a). Outline of MIIAD. MIIAD processes modallity-incomplete data, generating anomaly scores and segmentation maps. Our model \textbf{RADAR} surpasses most MIAD methods on this task. (b). Overfitting from modality-incomplete training. Training with incomplete-modal data can cause the model to focus on irrelevant features, leading to overfitting.}
\label{fig:intro}
\vspace{-15pt}
\end{figure}

Industrial Anomaly Detection (IAD) is pivotal in industrial quality inspection, focusing on product defects~\cite{gudovskiy2022cflow, roth2022cvpr}. Recent studies~\cite{wang2023multimodal} emphasize 3D geometries and color characteristics in defect identification, suggesting systems integrating 3D point clouds with 2D images represent a promising method. In recent years, emerging multimodal Transformer-based methods~\cite{weiuninet2025, xu2025zeroshotanomalydetectionreasoning} provide solutions for Multimodal Industrial Anomaly Detection (MIAD).

However, previous MIAD work typically assumes complete data availability~\cite{ma2022multimodal}, an assumption rarely feasible in industrial environments due to equipment limitations and privacy concerns, particularly with 3D data acquisition~\cite{sui2024incompletemultimodalindustrialanomaly}. Recent studies~\cite{Wei_2023_CVPR, lee2023cvpr} have emphasized that exploring the issue of modality-incompleteness, including corresponding benchmark and solution research, is meaningful and practical. To bridge the gap, we first introduce a novel and challenging task: \textbf{\underline{M}odality-\underline{I}ncomplete \underline{I}ndustrial \underline{A}nomaly \underline{D}etection (MIIAD)}. And to support this, we propose a comprehensive benchmark, \textbf{\textit{MIIAD Bench}}, which collects multimodal industrial anomaly datasets from MVTec-3D AD~\cite{bergmann2022mvtec} and Eyecandies~\cite{Bonfiglioli_2022_ACCV} datasets, and constructs primary 30\%, 50\%, and 70\% modality-missing data splits under the guidance of experts in industrial anomaly detection. This benchmark aims to simulate various missing-modality scenarios may occur in data sample, such as accessing only 2D RGB images or 3D point clouds during training or testing phases. To our knowledge, we are the first to systematically propose the MIIAD task and provide a benchmark with rich, expert-guided data splits. We look forward to contributing to the advancement of MIAD.

With our new benchmark, we evaluate state-of-the-art (SOTA) MIAD models and identify key limitations: \textbf{i) Lack of robustness to varying modality-missing scenarios:} Unsurprisingly, as shown in Fig.~\ref{fig:intro}~a, SOTA models suffer sharp performance drops, revealing struggles with MIIAD. Modality absence (at any rate, training/testing) creates diverse incomplete scenarios~\cite{lee2023cvpr}, yet even advanced missing-modality methods~\cite{sui2024incompletemultimodalindustrialanomaly, wang2023multi} fail to boost MIAD performance on \textbf{\textit{MIIAD Bench}}. This highlights urgent need for robust IAD methods adaptable to incomplete modalities. \textbf{ii) Overfitting from training on modality-incomplete data:} Additionally, training on such data risks overfitting~\cite{NEURIPS2023_abb4847b}; e.g., single-modal data struggles to support some tasks, leading models to overfit irrelevant features. As in Fig.~\ref{fig:intro}~b, models learning from anomaly-unidentifiable 2D data focus more on irrelevant features, causing overfitting. This demands methods robust to distinguishing different modality combinations and mitigating overfitting.

To facilitate robust detection in such an imperfect environment, we propose \textbf{RADAR}, a \textbf{\underline{R}}obust mod\textbf{\underline{A}}lity-instructive fusing and \textbf{\underline{D}}etecting fr\textbf{\underline{A}}mewo\textbf{\underline{R}}k. It enhances the model's ability to handle incomplete modalities in both the feature fusion and anomaly detection stages, comprising two main components: 
\textbf{i) Stage1: Adaptive Instruction Fusion:} In the feature fusion, we treat modality-complete and modality-incomplete scenarios as different inputs, constructing modality-incomplete instructions based on prompt-learning. These instructions guide the multimodal Transformer to robustly adapt to various modality-incomplete scenarios through trainable parameters. We also propose a Adaptive Learning Module based on HyperNetwork to dynamically adjust the parameters of the multimodal Transformer. 
\textbf{ii) Stage2: Double-Pseudo Hybrid Detection:} To mitigate overfitting issues caused by missing modality data, we design a Double-Pseudo Hybrid Module in the anomaly detection. This module employs a double-head structure with different supervision to highlight the uniqueness of different modality combinations, enabling the model to better focus on features that can complete the anomaly detection task. Finally, we integrate features stored in three repositories and make decisions based on the Mahalanobis Distance Matrix (MDM) and One-Class Support Vector Machine (OCSVM).

% \begin{table}[!t]
%     \begin{center}
%     \caption{\textbf{\textit{MIIAD Bench} - Eyecandies Split with different missing types and rates.}}
%     \vspace{-10pt}
%     \captionsetup{font={small,stretch=1.25}, labelfont={bf}}
%      \renewcommand{\arraystretch}{1.1}
%      \resizebox{0.48\textwidth}{!}{
%       {\small\begin{tabular}{c|c|c|c|c|c}
%        \toprule[1.5pt]
%        \textbf{Missing Rate} & \textbf{Type} & \textbf{\makecell{3D PC\\in \# Train}} & \textbf{\makecell{2D RGB\\in \# Train}} & \textbf{\makecell{3D PC\\in \# Test}} & \textbf{\makecell{2D RGB\\in \# Test}} \\
%        \hline
%        \multirow{3}{*}{30\%} & 3D PC & 7000 & 10000 & 2800 & 4000 \\
%        & 2D RGB & 10000 & 7000 & 4000 & 2800 \\
%        & Both & 8500 & 8500 & 3400 & 3400 \\
%        \hline
%        \multirow{3}{*}{50\%} & 3D PC & 5000 & 10000 & 2000 & 4000 \\
%        & 2D RGB & 10000 & 5000 & 4000 & 2000 \\
%        & Both & 7500  & 7500  & 3000  & 3000  \\
%        \hline
%        \multirow{3}{*}{70\%} & 3D PC & 3000 & 10000 & 1200 & 4000 \\
%        & 2D RGB & 10000 & 3000 & 4000 & 1200 \\
%        & Both & 6500  & 6500 & 2600 & 2600 \\
%        \toprule[1.5pt]
%       \end{tabular}}
%       }
%       \label{tab:data_split2}
%     \end{center}
%     \vspace{-21pt}
% \end{table}

Extensive experiments demonstrate broad value of our work: \textbf{i) Introducing a challenging practical new task:} Recent advanced MIAD methods struggle significantly on our proposed \textbf{\textit{MIIAD Bench}}. For instance, Shape-Guided~\cite{pmlr-v202-chu23b} M3DM~\cite{wang2023multimodal} show at least $3\%$ metric degradation in the $30\%$ split compared to the modality-complete setting, and at least $16\%$ degradation at $70\%$ missing rate. Meanwhile, missing-modality based methods provide merely ~$1\%$ improvement over baselines. This confirms MIIAD's substantial challenge practical value. \textbf{ii) Improving existing advanced MIAD methods:} Conversely, our \textbf{RADAR} model achieves SOTA performance on \textbf{\textit{MIIAD Bench}}, outperforming the M3DM baseline by at least $3\%$ and $5\%$ at $30\%$ and $70\%$ missing rates respectively across all metrics. This validates our two-stage framework's effectiveness. \textbf{iii) Outperforming other missing-modality based methods:} We significantly surpass missing-modality approaches, achieving at least $4\%$ higher metrics at $70\%$ missing rate, demonstrating our framework's broad applicability for modality-missing scenarios.

Our contributions are threefold:
\vspace{-4pt}

\begin{itemize}
    \item  We propose a new challenging and practical task: \textbf{\underline{M}}odality-\textbf{\underline{I}}ncomplete \textbf{\underline{I}}ndustrial \textbf{\underline{A}}nomaly \textbf{\underline{D}}etection (\textbf{MIIAD}), and design a new benchmark, \textbf{\textit{MIIAD Bench}}.  
    \item  We propose an end-to-end MIIAD learning framework, called \textbf{RADAR}, which enhances the robustness of the multimodal Transformer in a modality-incomplete setting.  
    \item Experiments demonstrate that, through modality-incomplete instruction and Double-Pseudo structure training, our framework significantly outperforms other methods in MIIAD.
\end{itemize}

\vspace{-7pt}
\section{Related Work}
\label{sec:related_work}
\newcommand{\red}[1]{\textcolor{red}{#1}}

\vspace{-3pt}

\subsection{Industrial Anomaly Detection}

\textbf{2D Industrial Anomaly Detection} can be divided into: 
i) \textit{Feature Extraction Methods}: Approaches such as teacher-student architecture~\cite{bergmann2020uninformed,salehi2021multiresolution} and one-class classification (OCC)~\cite{zhang2021anomaly,hu2021semantic} leverage pre-trained models or hypersphere-based strategies to distinguish anomalies. 
ii) \textit{Reconstruction-based Methods}: These use encoders and decoders to reconstruct images for pixel-level anomaly detection, including autoencoders~\cite{dehaene2020anomaly}, GANs~\cite{song2021anoseg,liang2023omni}, transformers~\cite{jiang2022masked}, and diffusion models~\cite{wyatt2022anoddpm}.

\textbf{3D Industrial Anomaly Detection} offers advantages in capturing spatial information beyond what is available in RGB images. The release of the MVTec 3D-AD dataset~\cite{bergmann2022beyond} has spurred several papers focusing on anomaly detection in 3D industrial images. Bergmann and Sattlegger~\cite{bergmann2023anomaly} introduced a teacher-student model for 3D anomaly detection, while Horwitz and Hoshen ~\cite{horwitz2023back}proposed a method that combines hand-crafted 3D representations with 2D features.

\vspace{-8pt}
\subsection{Modality-Incomplete Learning}
\vspace{-3pt}

Variational auto-encoders~\cite{van2018learning} effectively generate missing modalities unsupervised, but they typically overlook crucial shared vs. modality-specific feature distinctions~\cite{wang2023multi}. Robust-Mseg ~\cite{chen2019robust} addresses this by disentangling features into modality-specific appearance codes and modality-invariant content codes. SMIL~\cite{ma2021smil} achieves flexible incomplete data handling through Bayesian meta-learning, while ShaSpec~\cite{wang2023multi} offers a simpler yet effective architecture for multi-task learning. We follow these works' modality-missing settings to ensure method generality.

Compared with these methods with a single feature fusion stage, our method contains two stages to enhance the model's robustness.
% \vspace{-15pt}
\section{MIIAD}
\label{sec:MIIAD_Bench}

\newcommand{\MIIADDataset}{\mathcal{D}}

% \vspace{-2pt}
\subsection{Problem Formulation}

% \vspace{-3pt}
To systematically benchmark the robustness of current methods for industrial anomaly detection under modality-incomplete scenarios, we introduce a novel task called \textbf{\underline{M}odality-\underline{I}ncomplete \underline{I}ndustrial \underline{A}nomaly \underline{D}etection (MIIAD)}. MIIAD aims to assess model performance for industrial anomaly detection in diverse and complex modality-missing scenarios. We consider MIIAD tasks involving two modalities: 2D RGB image and 3D point cloud. Specifically, a MIIAD dataset, denoted as $\MIIADDataset$, can be formally divided into three subsets based on the modality missing situations: $\MIIADDataset = \{\MIIADDataset^C, \MIIADDataset^{2D}, \MIIADDataset^{3D}\}$. Here, $\MIIADDataset^C = \{(x_i^{2D}, x_i^{3D}, y_i)\}$ $(i \in {1, ..., |\MIIADDataset^C|})$ represents the modality-complete subset, $\MIIADDataset^{2D} = \{(x_i^{2D}, y_i)\}$ $(i \in {1, ..., |\MIIADDataset^{2D}|})$ comprises samples missing 3D point clouds, and $\MIIADDataset^{3D} = \{(x_i^{3D}, y_i)\}$ $(i \in {1, ..., |\MIIADDataset^{3D}|})$ includes samples lacking 2D RGB images. The primary task objective is to effectively perform IAD in various modality-incomplete settings.

% MVTec-3D AD~\cite{bergmann2022mvtec} dataset is the inaugural multimodal industrial anomaly detection dataset released by MVTec. It encompasses $2656$ training samples and $1137$ test samples, across 10 industrial categories. Alongside the 2D RGB image data within the dataset, high-resolution industrial 3D sensors are employed for depth scanning of products, capturing position information in 3-channel tensors that denote x, y, and z coordinates. Eyecandies~\cite{Bonfiglioli_2022_ACCV} dataset is a synthetic dataset with photo-realistic images of 10 candy categories for unsupervised anomaly detection. It provides RGB images, depth/normal maps, and auto-generated anomaly labels, eliminating human annotation bias.

\begin{table}[!t]
    \begin{center}
    \caption{\textbf{\textit{MIIAD Bench} - MVTec-3D AD Split with different missing types and rates.}}
    \vspace{-10pt}
    \captionsetup{font={small,stretch=1.25}, labelfont={bf}}
     \renewcommand{\arraystretch}{1.1}
     \resizebox{0.48\textwidth}{!}{
      {\small\begin{tabular}{c|c|c|c|c|c}
       \toprule[1.5pt]
       \textbf{Missing Rate} & \textbf{Type} & \textbf{\makecell{3D PC\\in \# Train}} & \textbf{\makecell{2D RGB\\in \# Train}} & \textbf{\makecell{3D PC\\in \# Test}} & \textbf{\makecell{2D RGB\\in \# Test}} \\
       \hline
       \multirow{3}{*}{30\%} & 3D PC & 1860 & 2656 & 840 & 1197 \\
       & 2D RGB & 2656 & 1860 & 1197 & 840 \\
       & Both & 2260 & 2260 & 1020 & 1020 \\
       \hline
       \multirow{3}{*}{50\%} & 3D PC & 1330 & 2656 & 600 & 1197 \\
       & 2D RGB & 2656 & 1330 & 1197 & 600 \\
       & Both & 1990 & 1990 & 900 & 900 \\
       \hline
       \multirow{3}{*}{70\%} & 3D PC & 800 & 2656 & 360 & 1197 \\
       & 2D RGB & 2656 & 800 & 1197 & 360 \\
       & Both & 1730 & 1730 & 780 & 780 \\
       \toprule[1.5pt]
      \end{tabular}}
      }
      \label{tab:data_split1}
    \end{center}
    \vspace{-20pt}
\end{table}

\vspace{-7pt}
\subsection{\textit{MIIAD Bench}}

% \vspace{-3pt}
\noindent$\textbf{Dataset Construction.}$ To cater to our research on modality-incom\\-plete in IAD, we reconstructed the two prevailing datasets MVTec-3D AD~\cite{bergmann2022mvtec} and Eyecandies~\cite{Bonfiglioli_2022_ACCV} into a modality-incomplete setting, called \textbf{\textit{MIIAD Bench}}, as shown in the upper part of Fig.~\ref{fig:model_structure}, guided by domain experts. In \textbf{\textit{MIIAD Bench}}, modality incompleteness may occur in either training or testing phases, with varying missing rates. The details of main data splits are shown in Tab.~\ref{tab:data_split1} and Tab.~~\ref{tab:data_split2}.

For a comprehensive study, we primarily set modality missing rates at $30\%$, $50\%$, and $70\%$ for three scenarios: 3D point cloud missing, RGB missing, and dual-modality missing. This aligns with prior work~\cite{10855484, Zhang_2022, wang2023multi, bao2022vlmo} on modality missing settings and is recognized by our consulted domain experts. We employed a random sampler to construct data covering three types of modality incompleteness: 3D point cloud modality-incomplete, 2D RGB image modality-incomplete, or both modalities-incomplete. For the first two scenarios, setting the missing rate of 3D point cloud data to $\eta\%$ implies the data comprises solely 2D RGB images of $\eta\%$ and complete data of $(1-\eta)\%$. Conversely, when the missing rate of 2D RGB image data is $\eta\%$, the situation is reversed. In cases where both modalities are incomplete with a missing rate of $\eta\%$, the data includes $\frac{\eta}{2}\%$ 2D RGB image data, $\frac{\eta}{2}\%$ 3D point cloud data, and $(1-\eta)\%$ complete data. We followed established modality-missing configurations and data processing protocols from recognized work~\cite{wang2023multi, bao2022vlmo} to ensure methodological soundness. 

% The detailed data split can be found in Tab.~\ref{tab:data_split1} (MVTec-3D AD) and Tab.~~\ref{tab:data_split2} (Eyecandies). 

\vspace{2pt}
\noindent$\textbf{Evaluation Metrics.}$ All evaluation metrics employed in our experiment align with those provided in the MVTec-3D AD~\cite{bergmann2022mvtec} and Eyecandies~\cite{Bonfiglioli_2022_ACCV} datasets. The image-level and pixel-level anomaly detection performance are evaluated with Area Under the Receiver Operating Curve (I-AUROC and P-AUROC, respectively), while the segmentation performance is assessed using Area Under the Per-Region Overlap (AUPRO).

\begin{table}[!t]
    \begin{center}
    \caption{\textbf{\textit{MIIAD Bench} - Eyecandies Split with different missing types and rates.}}
    \vspace{-10pt}
    \captionsetup{font={small,stretch=1.25}, labelfont={bf}}
     \renewcommand{\arraystretch}{1.1}
     \resizebox{0.48\textwidth}{!}{
      {\small\begin{tabular}{c|c|c|c|c|c}
       \toprule[1.5pt]
       \textbf{Missing Rate} & \textbf{Type} & \textbf{\makecell{3D PC\\in \# Train}} & \textbf{\makecell{2D RGB\\in \# Train}} & \textbf{\makecell{3D PC\\in \# Test}} & \textbf{\makecell{2D RGB\\in \# Test}} \\
       \hline
       \multirow{3}{*}{30\%} & 3D PC & 7000 & 10000 & 2800 & 4000 \\
       & 2D RGB & 10000 & 7000 & 4000 & 2800 \\
       & Both & 8500 & 8500 & 3400 & 3400 \\
       \hline
       \multirow{3}{*}{50\%} & 3D PC & 5000 & 10000 & 2000 & 4000 \\
       & 2D RGB & 10000 & 5000 & 4000 & 2000 \\
       & Both & 7500  & 7500  & 3000  & 3000  \\
       \hline
       \multirow{3}{*}{70\%} & 3D PC & 3000 & 10000 & 1200 & 4000 \\
       & 2D RGB & 10000 & 3000 & 4000 & 1200 \\
       & Both & 6500  & 6500 & 2600 & 2600 \\
       \toprule[1.5pt]
      \end{tabular}}
      }
      \label{tab:data_split2}
    \end{center}
    \vspace{-18pt}
\end{table}

% \vspace{-5pt}
\section{RADAR}
\label{sec:method}

\newcommand{\PointCloud}{\mathcal{P}}
\newcommand{\PointGroupFeature}{\mathcal{T}}

\begin{figure*}[t]
\includegraphics[width=0.99\textwidth]{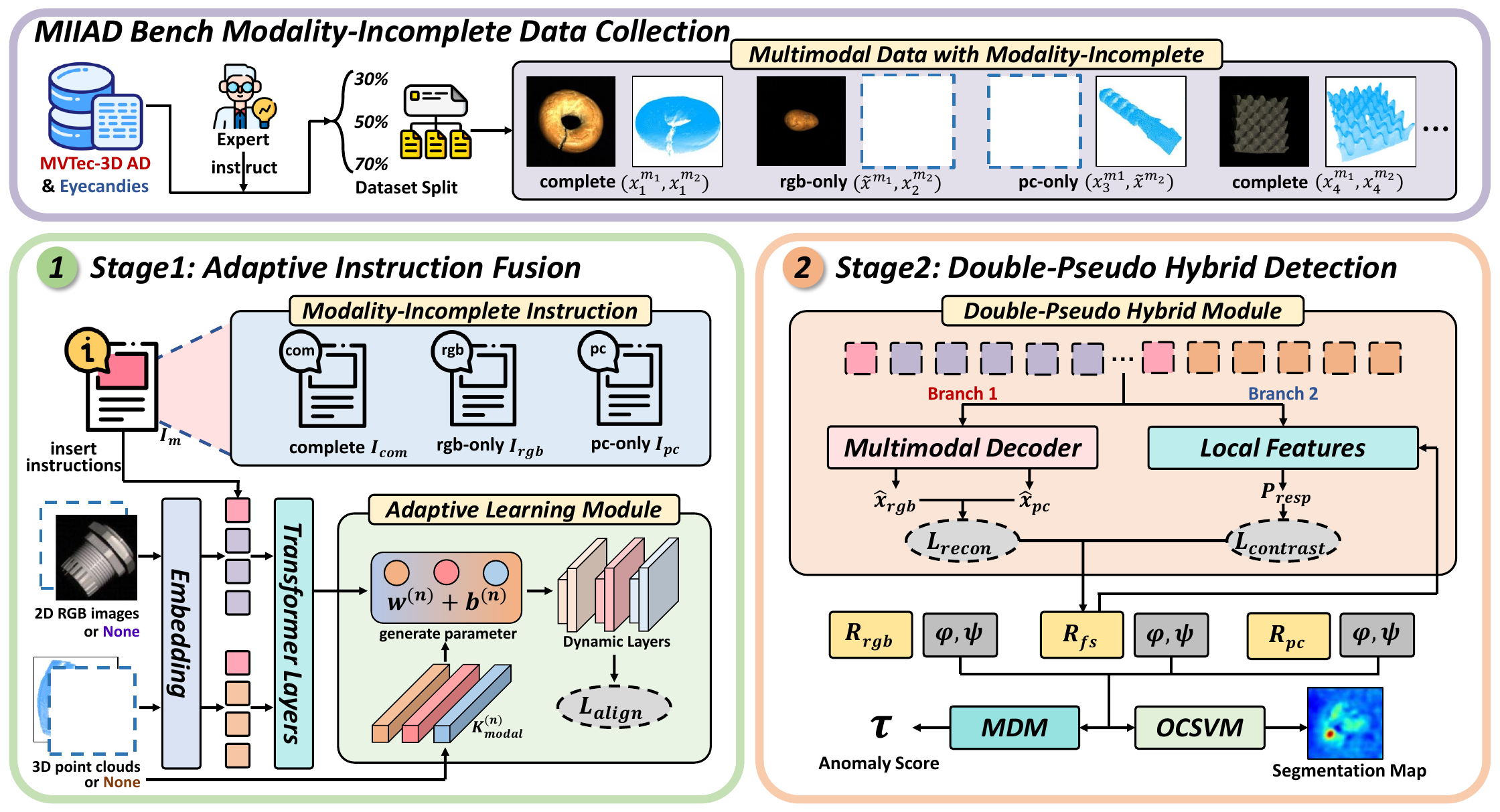}
\vspace{-12pt}
\centering\caption{Overview of \textbf{RADAR}. It consists of three parts: \textbf{i) \textit{MIIAD Bench} Data Collection.} Collecting data from the MVTec-3D AD dataset and constructing different data splits for the benchmark under the guidance of domain experts. \textbf{ii) Adaptive Instruction Fusion.} Prepending the modality-incomplete instruction $I_m$ of multimodal data $(x^{m_1}_i, x^{m_2}_i)$ into a simple multimodal Transformer input token, while introducing HyperNetwork to achieve adaptive parameter learning. \textbf{iii) Double-Pseudo Hybrid Detection.} Constructing a Double-Pseudo Hybrid Module for mitigating overfitting and feeding the features stored in multiple repositories $R_{rgb}$, $R_{fs}$, and $R_{pc}$ into the MDM and OCSVM for anomaly score $\tau$ and segmentation map, respectively. }  
\label{fig:model_structure}
\vspace{-12pt} 
\end{figure*}

This section introduces the \textbf{\underline{R}}obust mod\textbf{\underline{A}}lity-instructive fusing and \textbf{\underline{D}}etecting fr\textbf{\underline{A}}mewo\textbf{\underline{R}}k (\textbf{RADAR}). As shown in Fig.~\ref{fig:model_structure}, the \textbf{RADAR} framework addresses the practical MIIAD challenges through two key components: \textbf{i) Stage1: Adaptive Instruction Fusion (Section~\ref{subsec:method_stage1}):} It employs modality-incomplete instructions with inserted learnable parameters to guide the multimodal Transformer in adapting to modality-incomplete scenarios, while utilizing HyperNetwork for adaptive learning. \textbf{ii) Stage2: Double-Pseudo Hybrid Detection (Section~\ref{subsec:method_stage2}):} It designs a Double-Pseudo Hybrid Module with a double-head structure to highlight the uniqueness of different modality combinations, and makes decisions based on MDM and OCSVM to mitigate overfitting issues.

\vspace{-3pt}
\subsection{Pre: 2D\&3D Feature Extraction}
\label{subsec:method_pre}

If the modality data is incomplete, we use corresponding pseudo-inputs as substitutes: for 2D images, we employ a full-one pixel value image as the pseudo-input $\tilde{x}^{m1}$; for 3D point clouds, we generate a full-one tensor as the pseudo-input $\tilde{x}^{m2}$ with dimensions matching the read data size. When the modality data is complete, we extract features using the following approach.

\vspace{3pt}
\noindent$\textbf{2D Images Feature Extraction.}$ We employ pretrained ViT~\cite{dosovitskiy2021image} from HuggingFace~\cite{wolf-etal-2020-transformers} to extract 2D image features $f_{rgb}$, consistent with previous research. Due to space constraints, further details will not be provided.

\vspace{3pt}
\noindent$\textbf{3D Point Clouds Feature Extraction.}$  
The input point cloud $\PointCloud$ is an unordered sequence of point positions in three-dimensional space, containing $N_{\PointCloud}$ points, which makes it challenging to extract meaningful features directly by feeding them into standard Transformer modules. We first divid the point cloud $\PointCloud$ into $M$ groups $\PointCloud = \{\PointCloud_1, ..., \PointCloud_M\}$ using Farthest Point Sampling (FPS)~\cite{NIPS2017_d8bf84be}. For each group $\PointCloud_i$, we further extract feature vector $\PointGroupFeature_i$ utilizing a pretrained MaskTransformer~\cite{yu2022pointbert} and Point-MAE~\cite{pang2022masked}, \emph{i.e.}, each group $\PointCloud_i$ has a single point feature $\PointGroupFeature_i$. We take the output of the MaskTransformer as the initial point features $\PointGroupFeature_{en} = [{\PointGroupFeature_{en}}_{1}, ..., {\PointGroupFeature_{en}}_{M}]$. $\PointGroupFeature_{en}$ will be fed to Point-MAE to further improve the 3D point feature and take the decoder's output as the extracted 3D point features $\PointGroupFeature = [\PointGroupFeature_1, ..., \PointGroupFeature_M]$, which can be denoted as $\PointGroupFeature=Decoder_{MAE}(\PointGroupFeature_{en})$.

After point interpolation and projection, we apply average pooling to derive 3D point cloud features $f_{\text{pc}}$.

% \noindent$\textbf{Point Interpolation and Projection.}$  To prevent the imbalanced feature density issue caused by FPS, we interpolate the features back into the original point cloud, which can be denoted as
% \begin{equation}
% \begin{aligned}
% p'_j=\sum^M_{i=1}{\alpha_i \PointGroupFeature_i}\,,\,\,
% \alpha_i=\frac{\frac{1}{||c_i-p_j||_2+\epsilon}}{\sum^M_{k=1}\sum^{N_{\PointCloud}}_{t=1}\frac{1}{||c_k-p_t||_2+\epsilon}}\,,
% \end{aligned}
% \end{equation}
% where $p_j, j\in \{1, 2, ..., N_{\PointCloud}\}$ represents a point in the original point cloud data, $p'_j, j\in \{1, 2, ..., N_{\PointCloud}\}$ denotes the point after interpolation, $c_i$ is the center point associated with the corresponding group $\PointCloud_i$, and $\epsilon$ is a fairly small constant. Subsequently, we apply average pooling and additional operations to derive 3D point cloud features $f_{\text{pc}}$.

\vspace{-15pt}
\subsection{Stage1: Adaptive Instruction Fusion}
\label{subsec:method_stage1}

For modality-incomplete scenarios, we maintain that focusing on existing modality features during fusion outweighs generating missing data. Inspired by prompt learning's prefix-tuning~\cite{lee2023cvpr}, at Stage1, we construct modality-incomplete instructions (as learnable parameters) and insert them into the multimodal Transformer for feature fusion. And then propose an Adaptive Learning Module based on HypernetWork for parameter optimization.

\vspace{2pt}
\noindent$\textbf{Modality-Incomplete Instruction.}$ The distinct inputs between complete and incomplete modality data guide our learnable modality-incomplete instructions~\cite{lee2023cvpr}. Thus, we create \textbf{modality-incomplete instructions} for each modality-incomplete scenario. These instructions are implemented as randomly initialized parameters. 

As shown in Fig.~\ref{fig:model_structure}, for IAD tasks involving two modalities, we configure $2^2-1=3$ different instructions, which are then prepended to the input tokens according to the modality-incomplete conditions. Only the instruction corresponding to a specific missing modality condition will be activated (i.e., trained). The input embedding feature of the $j$-th Multi-head Self-Attention (MSA) layer is represented as $h^j\in \mathbb R^{L_i\times d},i=1,2,...,N$, where $L_i$ is the instruction length and $d$ is the embedding dimension. The function $F_{instruct}$ describing the instruction addition process can be expressed as:
\begin{equation}
F_{instruct}(i^j_m,h^j)=\text{concat}(i^j_m;h^j)\,,
\end{equation}

\noindent$\textbf{Instructed Feature Fusion.}$ In \textbf{RADAR}, the multimodal Transformer simply prepends the modality-incomplete instruction $i^j_m$ to the previously extracted 2D feature $f_{rgb}$ and 3D feature $f_{pc}$ at the $j$-th layer respectively. Thus, we form the modified embedding:
\begin{equation}
{\hat{f_{ex}}}^j_i=F_{instruct}(f_m,h^j),\,\,f_m\in\{f_{rgb},f_{pc}\}\,,
\end{equation}
The two new modified features can be represented as $\{\hat f_{pc},\hat f_{rgb}\}$.

Subsequently, through Multi-Layer Perceptrons (MLPs) $MLP_{pc}$ and $MLP_{rgb}$ along with fully connected layers, the features of both modalities are mapped to $\{g_{pc}, g_{rgb}\}$. The final fusion learning uses patch-wise contrastive loss (InfoNCE Loss~\cite{Oord2018RepresentationLW}) as the objective function, expressed as:
\begin{equation}
L_{align}=\frac{g_{pc}\cdot g^T_{rgb}}{\sum^{N_b}_{t=1}\sum^{N_p}_{k=1}g_{pc}\cdot g^T_{rgb}},
\end{equation}
where $N_b$ is the batch size and $N_p$ is the number of patches. The multimodal feature output from fully connected layer is called $\hat G$.

During training, the modality-incomplete instructions guide the model to learn features from the non-missing modality, enabling the model to better focus on the available effective feature information and achieve instructed feature fusion.

\vspace{5pt}
\noindent$\textbf{Adaptive Learning Module.}$ Recent work~\cite{ref:hypernetworks} demonstrates HyperNetworks' effectiveness for domain adaptation~\cite{liu2025boostingprivatedomainunderstanding} tasks like modality-incomplete scenarios, as it can process data from different sources as a single source domain without requiring perfect feature alignment. Thus, we propose an Adaptive Learning Module to leverage HyperNetwork for adaptive parameter learning of learnable components (treated as ``dynamic layers'') in Stage1.

% Recent work~\cite{ref:hypernetworks} demonstrates HyperNetworks' effectiveness for domain adaptation tasks like modality incompleteness. We propose an Adaptive Learning Module based on HyperNetwork that both generates parameters for network optimization and adaptively learns Stage1's dynamic layer components, better addressing MIIAD challenges.

Specifically, the HyperNetworks considers the parameters of the MLP layers as matrices $K^{(n)}\in\mathbb R^{N_{in}\times N_{out}}$, where $N_{in}$ and $N_{out}$ represent the number of input and output neurons in the $n$-th layer of the MLP, respectively. The creation process of $K^{(n)}$ can be viewed as matrix factorization, expressed as:
\begin{equation}
K^{(n)}=\xi (z^{(n)};\Theta_p),
\end{equation}
where both $z^{(n)}$ and $\xi(\cdot)$ are randomly initialized during training with parameters $\Theta_p$. Different from conventional HyperNetworks, to adapt to different modality-missing conditions, we propose modeling the parameters by replacing $z^{(n)}$ with input features representing modality-missing conditions. Specifically, a layer-specific encoder $\xi^h(\cdot)$ encodes $\{\hat f_{pc},\hat f_{rgb}\}$ into $e^{(n)}$. Then the HyperNetwork transforms $e^{(n)}$ into parameters, i.e., we input $e^{(n)}$ into the following two MLP layers to generate adaptive parameters:
\begin{equation}
w^{(n)}=(W_1\xi^h(e^{(n)}))+B_1)W_2+B_2,
\end{equation}
where $W_1$, $W_2$, $B_1$, and $B_2$ are the weights and bias terms of the first and second MLP layers. The model will adaptively adjust parameters based on different inputs, thereby enhancing robustness.

\subsection{Stage2: Double-Pseudo Hybrid Detection}
\label{subsec:method_stage2}

In stage2, since we used modality-incomplete instructions in stage1 to make the model focus more on existing modality features, this naturally leads to overfitting risks. For instance, when only 2D image data exists, the single-viewpoint 2D image may miss anomalies that would be visible in the absent 3D point cloud data. Requiring the model to predict anomalies using just 2D images could make it overfit to irrelevant features like lighting or textures~\cite{NEURIPS2023_abb4847b}.

To address this, we propose a Double-Pseudo Hybrid Module using pseudo-supervision to highlight modality-specific characteristics, followed by MDM and OCSVM-based anomaly detection using three feature repositories.

\vspace{5pt}
\noindent$\textbf{Double-Pseudo Hybrid Module.}$ To mitigate overfitting, we employ two distinct supervisory signals in two branches to optimize the multimodal feature $\hat G$ obtained from stage1, enabling it to learn diverse semantic information. Since MIAD is an unsupervised learning task, we construct two different pseudo-supervisions as follows:

For \textbf{Branch 1}, we impose global consistency constraints on $\hat G$ through self-supervised reconstruction. Specifically, we introduce a multimodal decoder comprising a 2D RGB decoder and a 3D point cloud decoder. $\hat G$ is fed into the decoder to obtain reconstructed RGB images $\hat{x}_{\text{rgb}}$ and point clouds $\hat{x}_{\text{pc}}$, then we compute the reconstruction loss using Mean Square Error (MSE):
\begin{equation}
\mathcal{L}_{\text{recon}} = \lambda_{\text{rgb}} \cdot \text{MSE}(x_{\text{rgb}}, \hat{x}_{\text{rgb}}) + \lambda_{\text{pc}} \cdot \text{MSE}(x_{\text{pc}}, \hat{x}_{\text{pc}})
\end{equation}
where $\lambda_{\text{rgb}} = 1.0$ and $\lambda_{\text{pc}} = 0.5$ are weighting factors, and $x_{\text{rgb}}$ and $x_{\text{pc}}$ the original multimodal inputs. A detailed discussion of the weighting strategy is in Appendix Sec.~9.

% This minimizes reconstruction error, forcing model to preserve global semantic information.

For \textbf{Branch 2}, we provide local discriminative constraints for $\hat G$ through contrastive learning. Specifically, we partition $\hat G$ into local patches by spatial location and compress them via average pooling into feature vectors $\{\mathbf{g}_1, \mathbf{g}_2, ..., \mathbf{g}_M\}$ as local features. Simultaneously, we maintain a multimodal online feature bank $R_{fs}$, initially randomly initialized and gradually replaced with real features during training. For each local patch $\mathbf{g}_i$, we compute cosine similarity with all features $\mathbf{f}^{\text{Memory}}$ in the bank and take the mean of top-K similarities as normality score $s_i$ and pseudo anomaly score $a_i$:
\begin{equation}
s_i = \frac{1}{K} \sum_{k=1}^K \text{sim}(\mathbf{g}_i, \mathbf{f}_k^{\text{Memory}})
\end{equation}
\begin{equation}
a_i = 1 - s_i
\end{equation}
Based on $a_i$, we get a low-resolution response map $\mathbf{P}_{\text{resp}}$ by spatial reconstruction and obtain pseudo-labels $y_p$ by thresholding: \(
y_{p}=\begin{cases} 
1, & \text{if}\,\,a_i>\sigma \\ 
0, & \text{otherwise} 
\end{cases},
\) where $\sigma = 0.35$ is a threshold. Finally, the contrastive loss is computed via Binary Cross-Entropy (BCE):
\begin{equation}
\mathcal{L}_{\text{contrast}} = \text{BCE}(\mathbf{P}_{\text{resp}}, y_p)
\end{equation}

This double-branch design constrains model learn globally consistent features (avoiding sensitivity local noise) while focusing anomalous regions (preventing overfitting normal samples).

\vspace{5pt}
\noindent$\textbf{Repository-based Anomaly Detection.}$ Subsequently, we use multiple repositories to store features. The 3D point cloud features, 2D RGB image features, and modality fusion features will be stored in repositories $R_{pc}$, $R_{rgb}$, and $R_{fs}$ respectively. Each repository is capable of generating predicted anomaly scores and segmentation maps. Mahalanobis Distance Matrix (MDM) is used to predict the final anomaly score $\tau$ and OCSVM is used to predict the final segmentation map $Seg_m$:
\begin{equation}
{\tau}=MDM(\{\phi(R_{pc},f_{pc}),\phi(R_{rgb},f_{rgb}),\phi(R_{fs},f_{fs})\}),
\end{equation}
\begin{equation}
{Seg_m}=OCSVM(\{\psi(R_{pc},f_{pc}),\psi(R_{rgb},f_{rgb}),\psi(R_{fs},f_{fs})\}),
\end{equation}
where $\phi,\psi$ are the score functions introduced in PatchCore~\cite{Roth_2022_CVPR}.

\begin{table*}[t]
  \begin{center}
  \captionsetup{font={small,stretch=1.25}, labelfont={bf}}
  \caption{Performance comparison of different MIAD methods on \textbf{\textit{MIIAD Bench}}. \textcolor{deepred}{\textbf{Red bold values}} indicate the best experimental performance, while \underline{underlined values} denote the second-best results.}
  \vspace{-10pt}
   \renewcommand{\arraystretch}{1.2}
   \resizebox{1\textwidth}{!}{
    \begin{tabular}{c|c||c c||c c||c c c c c||c c c c c}
     \toprule[1.5pt]
     
      \multirow{2}{*}[-0.5ex]{\makecell[c]{\textbf{Missing}\\\textbf{Rate}}} &
      \multirow{2}{*}[-0.5ex]{\makecell[c]{\textbf{\textit{MIIAD}}\\\textbf{\textit{Bench}}\\\textbf{Split}}}  & 
      \multicolumn{2}{c||}{\textbf{Data in \# Train}} & 
      \multicolumn{2}{c||}{\textbf{Data in \# Test}} & 
      \multicolumn{5}{c||}{\textbf{I-AUROC}} &
      \multicolumn{5}{c}{\textbf{AUPRO}} \\
      \cline{3-16}
      & & \textbf{3D PC} & \textbf{2D RGB} & \textbf{3D PC} & \textbf{2D RGB} & 
      \textbf{GPT-4V} & \makecell[c]{\textbf{Anomaly}\\\textbf{-GPT}~\cite{gu2023anomalyagpt}} & \textbf{M3DM}~\cite{wang2023multimodal} & \makecell[c]{\textbf{Shape-G}\\\textbf{-uided}~\cite{pmlr-v202-chu23b}} & \makecell[c]{\textbf{RADAR}\\\textbf{(Ours)}} &
      \textbf{Gemini-V} & \textbf{GPT-4V} & \textbf{M3DM}~\cite{wang2023multimodal} & \makecell[c]{\textbf{Shape-G}\\\textbf{-uided}~\cite{pmlr-v202-chu23b}} & \makecell[c]{\textbf{RADAR}\\\textbf{(Ours)}} \\\hline\hline

      \multirow{2}{*}{0\%}
      & \makecell[c]{MVTec-\\3D AD} & \multicolumn{2}{c||}{100\%} & \multicolumn{2}{c||}{100\%} & 0.753 & 0.922 & 0.945 & \textcolor{deepred}{\textbf{0.947}} & \textcolor{deepred}{\textbf{0.947}} & 0.344 & 0.572 & 0.964 & \textcolor{deepred}{\textbf{0.976}} & \underline{0.967} \\\cline{2-16}
      & \makecell[c]{Eye-\\candies} & \multicolumn{2}{c||}{100\%} & \multicolumn{2}{c||}{100\%} & 0.676 & 0.857 & \underline{0.897} & 0.891 & \textcolor{deepred}{\textbf{0.901}} & 0.297 & 0.514 & 0.882 & \underline{0.876} & \textcolor{deepred}{\textbf{0.885}} \\
      \hline

      \multirow{6}{*}{30\%} 
      & \multirow{3}{*}{\makecell[c]{MVTec-\\3D AD}}
      & 100\% & 70\% & 100\% & 70\% & 0.682 & 0.841 & 0.892 & \textcolor{deepred}{\textbf{0.910}} & \underline{0.908} & 0.315 & 0.544 & 0.911 & \underline{0.924} & \textcolor{deepred}{\textbf{0.925}} \\
      & & 70\% & 100\% & 70\% & 100\% & 0.693 & 0.828 & \underline{0.883} & 0.869 & \textcolor{deepred}{\textbf{0.903}} & 0.307 & 0.530 & \underline{0.932} & 0.918 & \textcolor{deepred}{\textbf{0.947}} \\
      & & 85\% & 85\% & 85\% & 85\% & 0.684 & 0.826 & \underline{0.886} & 0.883 & \textcolor{deepred}{\textbf{0.907}} & 0.309 & 0.540 & 0.917 & \underline{0.923} & \textcolor{deepred}{\textbf{0.929}} \\\cline{2-16}
      & \multirow{3}{*}{\makecell[c]{Eye-\\candies}}
      & 100\% & 70\% & 100\% & 70\% & 0.624 & 0.784 & 0.844 & \underline{0.864} & \textcolor{deepred}{\textbf{0.869}} & 0.264 & 0.453 & 0.828 & \underline{0.832} & \textcolor{deepred}{\textbf{0.847}} \\
      & & 70\% & 100\% & 70\% & 100\% & 0.615 & 0.788 & \underline{0.843} & 0.837 & \textcolor{deepred}{\textbf{0.858}} & 0.248 & 0.458 & \underline{0.832} & 0.829 & \textcolor{deepred}{\textbf{0.851}} \\
      & & 85\% & 85\% & 85\% & 85\% & 0.620 & 0.783 & 0.843 & \underline{0.850} & \textcolor{deepred}{\textbf{0.867}} & 0.252 & 0.454 & 0.830 & \underline{0.831} & \textcolor{deepred}{\textbf{0.854}} \\
      \hline

      \multirow{6}{*}{50\%} 
      & \multirow{3}{*}{\makecell[c]{MVTec-\\3D AD}}
      & 100\% & 50\% & 100\% & 50\% & 0.591 & 0.732 & 0.796 & \textcolor{deepred}{\textbf{0.811}} & \underline{0.810} & 0.248 & 0.485 & 0.835 & \underline{0.848} & \textcolor{deepred}{\textbf{0.857}} \\
      & & 50\% & 100\% & 50\% & 100\% & 0.584 & 0.724 & \underline{0.789} & 0.773 & \textcolor{deepred}{\textbf{0.821}} & 0.261 & 0.489 & \underline{0.849} & 0.838 & \textcolor{deepred}{\textbf{0.874}} \\
      & & 75\% & 75\% & 75\% & 75\% & 0.587 & 0.718 & 0.791 & \underline{0.799} & \textcolor{deepred}{\textbf{0.817}} & 0.253 & 0.483 & \underline{0.833} & 0.831 & \textcolor{deepred}{\textbf{0.862}} \\\cline{2-16}
      & \multirow{3}{*}{\makecell[c]{Eye-\\candies}}
      & 100\% & 50\% & 100\% & 50\% & 0.565 & 0.703 & 0.765 & \underline{0.786} & \textcolor{deepred}{\textbf{0.795}} & 0.202 & 0.387 & \underline{0.749} & 0.746 & \textcolor{deepred}{\textbf{0.772}} \\
      & & 50\% & 100\% & 50\% & 100\% & 0.559 & 0.695 & \underline{0.757} & 0.756 & \textcolor{deepred}{\textbf{0.780}} & 0.206 & 0.375 & \underline{0.741} & \underline{0.741} & \textcolor{deepred}{\textbf{0.769}} \\
      & & 75\% & 75\% & 75\% & 75\% & 0.554 & 0.701 & 0.759 & \underline{0.765} & \textcolor{deepred}{\textbf{0.787}} & 0.212 & 0.380 & \underline{0.742} & 0.738 & \textcolor{deepred}{\textbf{0.771}} \\
      \hline

      \multirow{6}{*}{70\%} 
      & \multirow{3}{*}{\makecell[c]{MVTec-\\3D AD}}
      & 100\% & 30\% & 100\% & 30\% & 0.476 & 0.616 & 0.675 & \underline{0.685} & \textcolor{deepred}{\textbf{0.707}} & 0.183 & 0.394 & 0.715 & \underline{0.729} & \textcolor{deepred}{\textbf{0.755}} \\
      & & 30\% & 100\% & 30\% & 100\% & 0.488 & 0.608 & \underline{0.662} & 0.649 & \textcolor{deepred}{\textbf{0.703}} & 0.184 & 0.401 & \underline{0.709} & 0.696 & \textcolor{deepred}{\textbf{0.748}} \\
      & & 65\% & 65\% & 65\% & 65\% & 0.484 & 0.615 & 0.668 & \underline{0.672} & \textcolor{deepred}{\textbf{0.706}} & 0.178 & 0.398 & 0.711 & \underline{0.714} & \textcolor{deepred}{\textbf{0.752}} \\\cline{2-16}
      & \multirow{3}{*}{\makecell[c]{Eye-\\candies}}
      & 100\% & 30\% & 100\% & 30\% & 0.437 & 0.571 & 0.629 & \underline{0.635} & \textcolor{deepred}{\textbf{0.662}} & 0.154 & 0.334 & 0.613 & \underline{0.616} & \textcolor{deepred}{\textbf{0.647}} \\
      & & 30\% & 100\% & 30\% & 100\% & 0.429 & 0.565 & \underline{0.623} & 0.611 & \textcolor{deepred}{\textbf{0.654}} & 0.162 & 0.328 & \underline{0.608} & 0.602 & \textcolor{deepred}{\textbf{0.649}} \\
      & & 65\% & 65\% & 65\% & 65\% & 0.433 & 0.570 & 0.624 & \underline{0.626} & \textcolor{deepred}{\textbf{0.658}} & 0.159 & 0.331 & 0.610 & \underline{0.615} & \textcolor{deepred}{\textbf{0.649}} \\
      
      \toprule[1.5pt]
    \end{tabular}
    }
    \label{tab:miiad_bench_with_MIAD}
  \end{center}
  \vspace{-15pt}
\end{table*}
\vspace{-3pt}
\section{Experiments}
\label{sec:experiments}

\subsection{Experimental Setup}

\noindent$\textbf{Baselines.}$ We conducted extensive comparative experiments with our \textbf{RADAR} model on \textbf{\textit{MIIAD Bench}}: \textbf{i) MIAD methods:} Recent advanced MIAD approaches including Shape-Guided~\cite{pmlr-v202-chu23b}, M3DM~\cite{wang2023multimodal}, and the Multimodal Large Language Models (MLLM) based AnomalyGPT~\cite{gu2023anomalyagpt}. Following the methodology of~\cite{xu2024custimizing}, we also incorporated powerful MLLM models Gemini Vision Pro and GPT-4 Vision into comparisons using specific prompts. \textbf{ii) Missing-modality based methods:} Recent advanced missing-modality approaches including ShaSpec~\cite{wang2023multi} using shared-specific feature modeling and CMDIAD~\cite{sui2024incompletemultimodalindustrialanomaly} with cross-modal distillation strategy. Additional comparative methods are presented in Appendix Sec.~2.

\begin{figure*}[!t]
\includegraphics[width=0.98\textwidth]{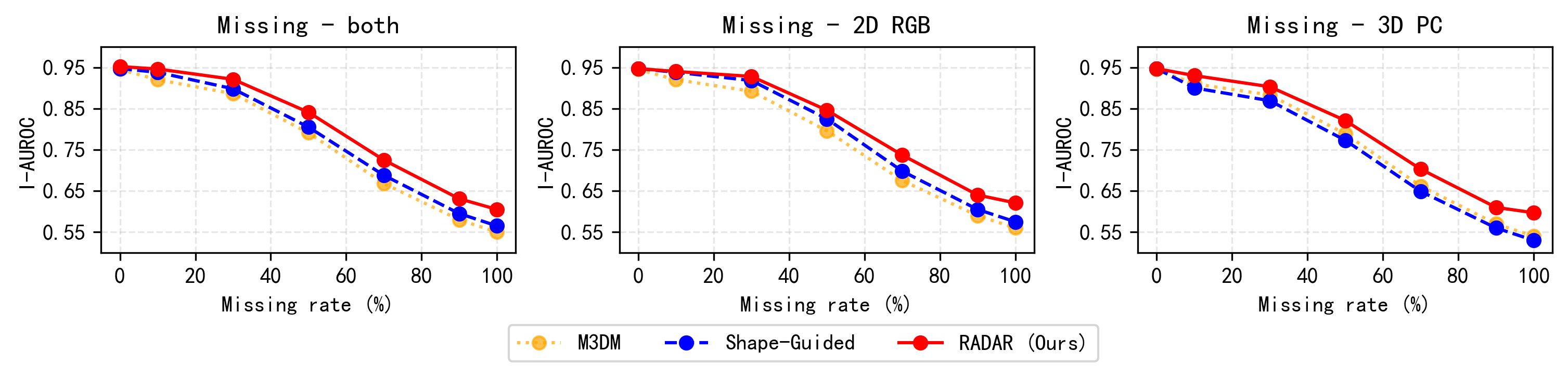}
\vspace{-12  pt}
\centering\caption{Quantitative results of I-AUROC on the \textit{MIIAD Bench} - MVTEC-3D AD with different missing rates under different missing-modality scenarios. Each data point in the figure represents that training and testing are with the same $\eta\%$ missing rate.} 
\label{fig:performance_under_different_modal_missing_rates}
\vspace{-10pt}
\end{figure*}

\vspace{3pt}
\noindent$\textbf{Implementation Details.}$ \textbf{RADAR}'s backbones consist of the following components: i) For 3D point cloud features, extraction is performed by MaskTransformer~\cite{yu2022pointbert} and Point-MAE Decoder~\cite{pang2022masked}. ii) For 2D RGB image features, the pre-trained ViTB/8 model with DINO is employed. Besides, we freeze all parameters of the multimodal Transformer backbone, focusing exclusively on training parameters relevant to the downstream MIIAD task. This approach significantly reduces computational costs during training. Further details of trainable components are provided in Appendix Sec.~1.

\vspace{-10pt}
\subsection{Results on \textit{MIIAD Bench}}

We compare \textbf{RADAR} with advanced MIAD models and missing-modality based methods on our \textbf{\textit{MIIAD Bench}}. Tab.~\ref{tab:miiad_bench_with_MIAD} shows the experimental results of I-AUROC and AUPRO metrics between \textbf{RADAR} and other MIAD models on \textbf{\textit{MIIAD Bench}}, while P-AUROC results are presented in Appendix Sec.~5 and Sec.~7. Tab.~\ref{tab:miiad_bench_with_imcomplete} demonstrates comparisons between \textbf{RADAR} and missing-modality based methods. Take the performance of the model on \textbf{\textit{MIIAD Bench}} - MVTec-3D AD Split as an example, the results reveal:

\textbf{i) We introduce a challenging and practical new task:} Tab.~\ref{tab:miiad_bench_with_MIAD} shows that the performance of advanced MIAD models degrades significantly with increasing modality missing rates. At $30\%$ 3D point cloud missing rate, Shape-Guided~\cite{pmlr-v202-chu23b} and M3DM~\cite{wang2023multimodal} show I-AUROC drops of $6.2$ points ($6.5\%$) and $7.8$ points ($8.2\%$), and AUPRO drops of $3.2$ points ($3.3\%$) and $5.8$ points ($5.9\%$) respectively compared to complete modality. Performance declines dramatically when missing rate exceeds $50\%$. For Shape-Guided~\cite{pmlr-v202-chu23b}, increasing 3D point cloud missing rate from $30\%$ to $50\%$ causes I-AUROC and AUPRO drops of $9.6$ points ($11.0\%$) and $8.0$ points ($8.7\%$); from $50\%$ to $70\%$, drops increase to $12.4$ points ($16.0\%$) and $14.2$ points ($16.9\%$). This demonstrates the significant challenge posed by MIIAD, which is unavoidable in real-world applications. Tab.~\ref{tab:miiad_bench_with_imcomplete} shows existing missing-modality methods struggle with MIIAD. At $70\%$ 3D point cloud missing rate, ShaSpec~\cite{wang2023multi} and CMDIAD~\cite{sui2024incompletemultimodalindustrialanomaly} only improve baseline I-AUROC by $0.9$ points ($1.3\%$) and $-1.0$ points ($-1.5\%$), and AUPRO by $0.9$ points ($1.2\%$) and $0.3$ points ($0.4\%$). Qualitative results in Fig~\ref{fig:performance_under_different_modal_missing_rates} confirm this trend across different missing modalities.

\begin{table*}[t]
  \begin{center}
  \captionsetup{font={small,stretch=1.25}, labelfont={bf}}
  \caption{Performance comparison with methods aiming at addressing the modality-incomplete issue on \textit{MIIAD Bench}.}
  \vspace{-10pt}
   \renewcommand{\arraystretch}{1.2}
   \resizebox{1\textwidth}{!}{
    \begin{tabular}{c|c||c c||c c||c c c c||c c c c}
     \toprule[1.5pt]
     
      \multirow{2}{*}[-0.5ex]{\makecell[c]{\textbf{Missing}\\\textbf{Rate}}} &
      \multirow{2}{*}[-0.5ex]{\makecell[c]{\textbf{\textit{MIIAD Bench}}\\\textbf{Split}}}  & 
      \multicolumn{2}{c||}{\textbf{Data in \# Train}} & 
      \multicolumn{2}{c||}{\textbf{Data in \# Test}} & 
      \multicolumn{4}{c||}{\textbf{I-AUROC}} &
      \multicolumn{4}{c}{\textbf{AUPRO}} \\
      \cline{3-14}
      & & \textbf{3D PC} & \textbf{2D RGB} & \textbf{3D PC} & \textbf{2D RGB} & 
      \makecell[c]{\textbf{M3DM}\\\textbf{(Baseline)}}~\cite{wang2023multimodal} & \makecell[c]{\textbf{+Sha}\\\textbf{-Spec}}~\cite{wang2023multi} & \textbf{+CMDIAD}~\cite{sui2024incompletemultimodalindustrialanomaly} & \makecell[c]{\textbf{+RADAR}\\\textbf{(Ours)}} &
      \makecell[c]{\textbf{M3DM}\\\textbf{(Baseline)}}~\cite{wang2023multimodal} & \makecell[c]{\textbf{+Sha}\\\textbf{-Spec}}~\cite{wang2023multi} & \textbf{+CMDIAD}~\cite{sui2024incompletemultimodalindustrialanomaly} & \makecell[c]{\textbf{+RADAR}\\\textbf{(Ours)}} \\\hline\hline

      \multirow{2}{*}{0\%} 
      & MVTec-3D AD & \multicolumn{2}{c||}{100\%} & \multicolumn{2}{c||}{100\%} & 0.945 & 0.943 & \textcolor{deepred}{\textbf{0.948}} & \underline{0.947} & \underline{0.964} & 0.958 & 0.960 & \textcolor{deepred}{\textbf{0.967}} \\\cline{2-14}
      & Eyecandies & \multicolumn{2}{c||}{100\%} & \multicolumn{2}{c||}{100\%} & 0.897 & 0.891 & \underline{0.898} & \textcolor{deepred}{\textbf{0.901}} & 0.882 & 0.882 & \underline{0.884} & \textcolor{deepred}{\textbf{0.885}} \\
      \hline

      \multirow{6}{*}{30\%} 
      & \multirow{3}{*}{\makecell[c]{MVTec-\\3D AD}}
      & 100\% & 70\% & 100\% & 70\% & 0.892 & \underline{0.901} & 0.898 & \textcolor{deepred}{\textbf{0.908}} & 0.911 & \underline{0.915} & 0.911 & \textcolor{deepred}{\textbf{0.925}} \\
      & & 70\% & 100\% & 70\% & 100\% & 0.883 & 0.887 & \underline{0.894} & \textcolor{deepred}{\textbf{0.903}} & 0.932 & \textcolor{deepred}{\textbf{0.938}} & 0.916 & \underline{0.947} \\
      & & 85\% & 85\% & 85\% & 85\% & 0.886 & 0.893 & \underline{0.897} & \textcolor{deepred}{\textbf{0.907}} & \underline{0.917} & \underline{0.917} & 0.913 & \textcolor{deepred}{\textbf{0.929}} \\\cline{2-14}
      & \multirow{3}{*}{\makecell[c]{Eye-\\candies}}
      & 100\% & 70\% & 100\% & 70\% & 0.844 & \underline{0.847} & 0.842 & \textcolor{deepred}{\textbf{0.869}} & 0.828 & \underline{0.834} & 0.831 & \textcolor{deepred}{\textbf{0.847}} \\
      & & 70\% & 100\% & 70\% & 100\% & 0.843 & \underline{0.851} & 0.845 & \textcolor{deepred}{\textbf{0.858}} & 0.832 & \underline{0.841} & 0.838 & \textcolor{deepred}{\textbf{0.851}} \\
      & & 85\% & 85\% & 85\% & 85\% & 0.843 & \underline{0.852} & 0.843 & \textcolor{deepred}{\textbf{0.867}} & 0.830 & \underline{0.840} & 0.835 & \textcolor{deepred}{\textbf{0.854}} \\
      \hline

      \multirow{6}{*}{70\%} 
      & \multirow{3}{*}{\makecell[c]{MVTec-\\3D AD}}
      & 100\% & 30\% & 100\% & 30\% & 0.675 & \underline{0.679} & 0.654 & \textcolor{deepred}{\textbf{0.707}} & 0.715 & \underline{0.717} & 0.703 & \textcolor{deepred}{\textbf{0.755}} \\
      & & 30\% & 100\% & 30\% & 100\% & 0.662 & \underline{0.671} & 0.652 & \textcolor{deepred}{\textbf{0.703}} & 0.709 & \underline{0.718} & 0.712 & \textcolor{deepred}{\textbf{0.748}} \\
      & & 65\% & 65\% & 65\% & 65\% & 0.668 & \underline{0.677} & 0.653 & \textcolor{deepred}{\textbf{0.706}} & 0.711 & \underline{0.717} & 0.707 & \textcolor{deepred}{\textbf{0.752}} \\\cline{2-14}
      & \multirow{3}{*}{\makecell[c]{Eye-\\candies}}
      & 100\% & 30\% & 100\% & 30\% & 0.629 & \underline{0.634} & 0.615 & \textcolor{deepred}{\textbf{0.662}} & 0.613 & \underline{0.614} & 0.606 & \textcolor{deepred}{\textbf{0.647}} \\
      & & 30\% & 100\% & 30\% & 100\% & 0.623 & \underline{0.624} & 0.607 & \textcolor{deepred}{\textbf{0.654}} & 0.608 & \underline{0.611} & 0.598 & \textcolor{deepred}{\textbf{0.649}} \\
      & & 65\% & 65\% & 65\% & 65\% & 0.624 & \underline{0.626} & 0.610 & \textcolor{deepred}{\textbf{0.658}} & 0.610 & \underline{0.613} & 0.601 & \textcolor{deepred}{\textbf{0.649}} \\
      \toprule[1.5pt]
    \end{tabular}
    }
    \label{tab:miiad_bench_with_imcomplete}
  \end{center}
  \vspace{-15pt}
\end{table*}

\begin{table}[t]
  \begin{center}
  \captionsetup{font={small,stretch=1.25}, labelfont={bf}}
  \caption{Ablation study of different modules under varying missing rates on \textit{MIIAD Bench} - MVTec-3D AD. Here, MII, ALM, and DPHM represent Modality-Incomplete Instruction, Adaptive Learning Module, and Double-Pseudo Hybrid Module.}
  \vspace{-10pt}
   \renewcommand{\arraystretch}{1.2}
   \resizebox{0.48\textwidth}{!}{
     \begin{tabular}{l|c c c||c c||c c}
      \toprule[1.5pt]
      
      \multirow{2}{*}{\textbf{Methods}} & \multicolumn{3}{c||}{\textbf{Setting}} & \multicolumn{2}{c||}{\textbf{50\% of pc}} & \multicolumn{2}{c}{\textbf{70\% of pc}}\\
      \cline{2-8}
      & MII & ALM & DPHM & I-AUROC & AUPRO & I-AUROC & AUPRO\\
      \hline \hline

      Baseline & & & & 0.789 & 0.849 & 0.662 & 0.709\\
      + MII & \faCheckCircle & & & 0.802 & 0.860 & 0.680 & 0.725\\
      + ALM & & \faCheckCircle & & 0.795 & 0.853 & 0.668 & 0.714\\
      + DPHM & & & \faCheckCircle & 0.798 & 0.856 & 0.675 & 0.719\\
      + MII \& ALM & \faCheckCircle & \faCheckCircle & & 0.810 & 0.865 & 0.688 & 0.732\\
      + MII \& DPHM & \faCheckCircle & & \faCheckCircle & \underline{0.816} & \underline{0.868} & \underline{0.693} & \underline{0.738}\\
      + ALM \& DPHM & & \faCheckCircle & \faCheckCircle & 0.804 & 0.862 & 0.682 & 0.727\\
      \textbf{RADAR (Ours)} & \faCheckCircle & \faCheckCircle & \faCheckCircle & \textcolor{deepred}{\textbf{0.821}} & \textcolor{deepred}{\textbf{0.874}} & \textcolor{deepred}{\textbf{0.703}} & \textcolor{deepred}{\textbf{0.748}}\\
      \toprule[1.5pt]
     \end{tabular}
     }
     \label{tab:ablation_study}
  \end{center}
  \vspace{-23pt}
\end{table}

\textbf{ii) Improving existing advanced MIAD methods:} As Tab.~\ref{tab:miiad_bench_with_MIAD} shows, our method excels in nearly all settings. Compared to M3DM baseline, we achieve I-AUROC improvements of $1.6$ points ($1.7\%$), $3.2$ points ($4.0\%$) and $4.1$ points ($6.2\%$) at $30\%$, $50\%$ and $70\%$ 3D point cloud missing rates respectively, with corresponding AUPRO gains of $1.5$ points ($3.9\%$), $2.5$ points ($2.9\%$) and $3.9$ points ($5.5\%$). Against Shape-Guided~\cite{pmlr-v202-chu23b} (which outperforms M3DM with complete modality), we achieve I-AUROC advantages of $3.4$ points ($4.0\%$), $4.8$ points ($6.2\%$) and $5.4$ points ($8.3\%$), and AUPRO advantages of $2.9$ points ($3.1\%$), $3.6$ points ($4.3\%$) and $5.2$ points ($7.4\%$) at respective missing rates. Fig~\ref{fig:performance_under_different_modal_missing_rates} shows our consistent performance gains as missing rates increase. Notably, our model shows increasingly larger performance gains over the baseline and competing MIAD methods as missing rates escalate, evidencing our framework's substantial improvements to current MIAD solutions.

\textbf{iii) Outperforming missing-modality based methods:} Tab.~\ref{tab:miiad_bench_with_imcomplete} demonstrates that using M3DM as the baseline, our approach shows stronger robustness over alternative modality-missing solutions. At $30\%$ 3D point cloud missing rate, we surpass CMDIAD~\cite{sui2024incompletemultimodalindustrialanomaly} and ShaSpec~\cite{wang2023multi} by $0.9$ points ($1.0\%$) and $3.6$ points ($4.0\%$) in I-AUROC, and $3.1$ points ($3.4\%$) and $0.9$ points ($0.9\%$) in AUPRO. At $70\%$ 3D point cloud missing rate, advantages increase to $5.1$ points ($7.8\%$) and $3.2$ points ($4.8\%$) in I-AUROC, and $3.6$ points ($5.0\%$) and $3.0$ points ($4.2\%$) in AUPRO. Notably, our model's performance gains over the baseline and other missing-modality methods increase progressively with higher missing rates.

% evidencing our framework's stronger robustness for MIIAD applications.

Similar patterns hold for 2D RGB missing case, both-modality missing case, and \textbf{\textit{MIIAD Bench}} - Eyecandies Split. More additional comparative experiments are provided in Appendix Sec.~6.

\begin{table}[!t]
  \begin{center}
  \captionsetup{font={small,stretch=1.25}, labelfont={bf}}
    \caption{Comparison with RGB-only and 3D-only methods on MVTec-3D AD. RADAR is trained with complete multimodal data but tested under 100\% modality absence scenarios.}
    \vspace{-10pt}
     \renewcommand{\arraystretch}{1.25}
     \resizebox{0.48\textwidth}{!}{
     
      \begin{tabular}{cc|c|c|c|c}
        \toprule
        \multicolumn{2}{c|}{\textbf{Methods}} & \textbf{Data in \# Test} & \textbf{I-AUROC} & \textbf{AUPRO} & \textbf{P-AUROC} \\
        \midrule
        
        \multicolumn{1}{c}{\multirow{5}{*}{\rotatebox{90}{RGB}}} 
        & PatchCore~\cite{Roth_2022_CVPR} & & 0.770 & 0.876 & 0.967 \\
        \multicolumn{1}{c}{} & CS-Flow~\cite{Gudovskiy_2022_WACV} & 100\% 2D RGB & 0.830 & 0.871 & / \\
        \multicolumn{1}{c}{} & CMDIAD~\cite{sui2024incompletemultimodalindustrialanomaly} & & \textcolor{deepred}{\textbf{0.858}} & \underline{0.943} & \underline{0.987} \\
        \multicolumn{1}{c}{} & M3DM~\cite{wang2023multimodal} & 0\% 3D PC & 0.850 & 0.942 & \underline{0.987} \\
        \multicolumn{1}{c}{} & \textbf{RADAR (Ours)} & & \underline{0.854} & \textcolor{deepred}{\textbf{0.944}} & \textcolor{deepred}{\textbf{0.989}} \\
        \midrule
        
        \multicolumn{1}{c}{\multirow{6}{*}{\rotatebox{90}{3D}}} 
        & FPFH~\cite{horwitz2022empirical} & & 0.782 & 0.924 & \underline{0.978} \\
        \multicolumn{1}{c}{} & Shape-Guided~\cite{pmlr-v202-chu23b} & 100\% 3D PC & \underline{0.916} & 0.931 & / \\
        \multicolumn{1}{c}{} & CMDIAD~\cite{sui2024incompletemultimodalindustrialanomaly} & & \textcolor{deepred}{\textbf{0.938}} & \textcolor{deepred}{\textbf{0.934}} & / \\
        \multicolumn{1}{c}{} & M3DM~\cite{wang2023multimodal} & 0\% 2D RGB & 0.874 & 0.906 & 0.970 \\
        \multicolumn{1}{c}{} & \textbf{RADAR (Ours)} & & 0.907 & \underline{0.932}  & \textcolor{deepred}{\textbf{0.984}}  \\
         
      \bottomrule
      
    \end{tabular}
  
    }
  \label{tab:compare_2D_3D}
  \end{center}
  \vspace{-15pt}
\end{table}

\begin{table*}[t]
  \begin{center}
  \captionsetup{font={small,stretch=1.25}, labelfont={bf}}
  \caption{Performance comparison of modality-incomplete in more situations.}
  \vspace{-10pt}
   \renewcommand{\arraystretch}{1.4}
   \resizebox{1\textwidth}{!}{
    \begin{tabular}{c|c||c c||c c||c c c c c||c c c c c}
     \toprule[1.5pt]
     
      \multirow{2}{*}[-0.5ex]{\makecell[c]{\textbf{Missing}\\\textbf{Rate}}} &
      \multirow{2}{*}[-0.5ex]{\makecell[c]{\textbf{\textit{MIIAD}}\\\textbf{\textit{Bench}}\\\textbf{Split}}}  & 
      \multicolumn{2}{c||}{\textbf{Data in \# Train}} & 
      \multicolumn{2}{c||}{\textbf{Data in \# Test}} & 
      \multicolumn{5}{c||}{\textbf{I-AUROC}} &
      \multicolumn{5}{c}{\textbf{AUPRO}} \\
      \cline{3-16}
      & & \textbf{3D PC} & \textbf{2D RGB} & \textbf{3D PC} & \textbf{2D RGB} & 
      \textbf{GPT-4V} & \makecell[c]{\textbf{Anomaly}\\\textbf{-GPT}~\cite{gu2023anomalyagpt}} & \textbf{M3DM}~\cite{wang2023multimodal} & \makecell[c]{\textbf{Shape-G}\\\textbf{-uided}~\cite{pmlr-v202-chu23b}} & \makecell[c]{\textbf{RADAR}\\\textbf{(Ours)}} &
      \textbf{Gemini-V} & \textbf{GPT-4V} & \textbf{M3DM}~\cite{wang2023multimodal} & \makecell[c]{\textbf{Shape-G}\\\textbf{-uided}~\cite{pmlr-v202-chu23b}} & \makecell[c]{\textbf{RADAR}\\\textbf{(Ours)}} \\\hline\hline

      \multirow{2}{*}{0\%}
      & \makecell[c]{MVTec-\\3D AD} & \multicolumn{2}{c||}{100\%} & \multicolumn{2}{c||}{100\%} & 0.753 & 0.922 & 0.945 & \textcolor{deepred}{\textbf{0.947}} & \textcolor{deepred}{\textbf{0.947}} & 0.342 & 0.574 & 0.964 & \textcolor{deepred}{\textbf{0.976}} & \underline{0.967} \\\cline{2-16}
      & \makecell[c]{Eye-\\candies} & \multicolumn{2}{c||}{100\%} & \multicolumn{2}{c||}{100\%} & 0.676 & 0.857 & \underline{0.897} & 0.891 & \textcolor{deepred}{\textbf{0.901}} & 0.297 & 0.514 & \underline{0.882} & 0.876 & \textcolor{deepred}{\textbf{0.885}} \\
      \hline

      \multirow{4}{*}{30\%} 
      & \multirow{2}{*}{\makecell[c]{MVTec-\\3D AD}}
      & \multicolumn{2}{c||}{100\%} & \multicolumn{2}{c||}{85\%} & 0.714 & 0.876 & 0.891 & \underline{0.897} & \textcolor{deepred}{\textbf{0.915}} & 0.320 & 0.543 & 0.924 & \underline{0.928} & \textcolor{deepred}{\textbf{0.934}} \\
      & & \multicolumn{2}{c||}{85\%} & \multicolumn{2}{c||}{100\%} & 0.706 & 0.858 & \underline{0.871} & 0.870 & \textcolor{deepred}{\textbf{0.895}} & 0.307 & 0.537 & 0.904 & \underline{0.916} & \textcolor{deepred}{\textbf{0.920}} \\\cline{2-16}
      & \multirow{2}{*}{\makecell[c]{Eye-\\candies}}
      & \multicolumn{2}{c||}{100\%} & \multicolumn{2}{c||}{85\%} & 0.651 & 0.816 & 0.849 & \underline{0.861} & \textcolor{deepred}{\textbf{0.877}} & 0.268 & 0.477 & 0.844 & \underline{0.852} & \textcolor{deepred}{\textbf{0.868}} \\
      & & \multicolumn{2}{c||}{85\%} & \multicolumn{2}{c||}{100\%} & 0.639 & 0.814 & 0.831 & \underline{0.837} & \textcolor{deepred}{\textbf{0.848}} & 0.251 & 0.463 & 0.814 & \underline{0.819} & \textcolor{deepred}{\textbf{0.843}} \\
      \hline

      \multirow{4}{*}{70\%} 
      & \multirow{2}{*}{\makecell[c]{MVTec-\\3D AD}}
      & \multicolumn{2}{c||}{100\%} & \multicolumn{2}{c||}{65\%} & 0.534 & 0.669 & 0.693 & \underline{0.698} & \textcolor{deepred}{\textbf{0.729}} & 0.183 & 0.417 & 0.732 & \underline{0.738} & \textcolor{deepred}{\textbf{0.780}} \\
      & & \multicolumn{2}{c||}{65\%} & \multicolumn{2}{c||}{100\%} & 0.516 & 0.644 & 0.644 & \underline{0.649} & \textcolor{deepred}{\textbf{0.675}} & 0.172 & 0.409 & 0.680 & \underline{0.688} & \textcolor{deepred}{\textbf{0.715}} \\\cline{2-16}
      & \multirow{2}{*}{\makecell[c]{Eye-\\candies}}
      & \multicolumn{2}{c||}{100\%} & \multicolumn{2}{c||}{65\%} & 0.489 & 0.603 & 0.629 & \underline{0.632} & \textcolor{deepred}{\textbf{0.674}} & 0.165 & 0.306 & 0.616 & \underline{0.624} & \textcolor{deepred}{\textbf{0.669}} \\
      & & \multicolumn{2}{c||}{65\%} & \multicolumn{2}{c||}{100\%} & 0.507 & 0.622 & \underline{0.601} & 0.598 & \textcolor{deepred}{\textbf{0.634}} & 0.164 & 0.297 & \underline{0.580} & 0.576 & \textcolor{deepred}{\textbf{0.623}} \\
      \toprule[1.5pt]
    \end{tabular}
    }
    \label{tab:radar_in_more_situations}
  \end{center}
  \vspace{-13pt}
\end{table*}

% \begin{table}[!ht]
%   \begin{center}
%   \captionsetup{font={small,stretch=1.25}, labelfont={bf}}
%     \caption{P-AUROC score results of Eyecandies dataset.}
%      \renewcommand{\arraystretch}{1.4}
%      \resizebox{0.48\textwidth}{!}{

%       \begin{tabular}{c|c|c|c}
%         \toprule
%         \textbf{Methods} & \makecell{Mean\\(30\% of pc)} & \makecell{Mean\\(50\% of pc)} & \makecell{Mean\\(70\% of pc)}\\
%         \midrule
%         {Voxel GAN}~\cite{bergmann2022mvtec} & 0.513 & 0.419 & 0.357\\
%         {Voxel VM}~\cite{bergmann2022mvtec} & 0.414 & 0.354 & 0.306\\
%         {PatchCore+FPFH}~\cite{horwitz2022empirical} & \colorbox{myyellow}{\textbf{0.747}} & \colorbox{myorange}{\textbf{0.669}} & \colorbox{myyellow}{\textbf{0.605}}\\
%         {AST}~\cite{Rudolph_2023_WACV} & 0.724 & 0.636 & 0.578\\
%         {M3DM}~\cite{wang2023multimodal} & \colorbox{myorange}{\textbf{0.754}} & \colorbox{myyellow}{\textbf{0.668}} & \colorbox{myorange}{\textbf{0.611}}\\
%         \textbf{\textbf{RADAR} (Ours)} & \colorbox{myred}{\textbf{0.796}} & \colorbox{myred}{\textbf{0.722}} & \colorbox{myred}{\textbf{0.674}}\\
%       \bottomrule
%       \end{tabular}

%     }
%   \label{tab:results_of_eyecandies_dataset}
%   \end{center}
%   \vspace{-0.5cm}
% \end{table}

\vspace{-2pt}
\subsection{Ablation Study} 
\vspace{2pt}

To evaluate effectiveness of each module in \textbf{RADAR}, we conducted comprehensive ablation studies on three key components: Modality-Incomplete Instruction (MII), Adaptive Learning Module (ALM), and Double-Pseudo Hybrid Module (DPHM). As shown in Tab.~\ref{tab:ablation_study}, each module significantly improves baseline performance. At $50\%$ 3D point cloud missing rate, these modules increase I-AUROC by $1.3$ points ($1.6\%$), $0.6$ points ($0.8\%$), and $0.9$ points ($1.1\%$), and AUPRO by $1.1$ points ($1.3\%$), $0.4$ points ($0.5\%$), and $0.7$ points ($0.8\%$) respectively. At $70\%$ missing rate, modules improve I-AUROC by $1.8$ points ($2.7\%$), $0.6$ points ($0.9\%$), and $1.3$ points ($1.9\%$), and AUPRO by $1.6$ points ($2.3\%$), $0.5$ points ($0.7\%$), and $1.0$ points ($1.4\%$).

The results reveal that the MII contributes most to performance gains, proving its effect for modality-missing scenarios. The DPHM shows secondary importance, reflecting its overfitting mitigation. Finally, the role of ALM shows that dynamic parameter learning enhances performance. Similar experimental patterns are observed for other modality-missing cases, as presented in Appendix Sec.~8.

% \begin{table}[!ht]
%   \begin{center}
%   \captionsetup{font={small,stretch=1.25}, labelfont={bf}}
%     \caption{More P-AUROC score results in different modality-incomplete situations.}
%     \vspace{-10pt}
%      \renewcommand{\arraystretch}{1.2}
%      \resizebox{0.48\textwidth}{!}{
     
%   \begin{tabular}{cccc}
%     \toprule
%     \textbf{Metrics} & M3DM (Baseline) & \textbf{\textbf{RADAR} (Ours)}\\
%     \midrule
%     Mean(modality-complete) & 0.926 & 0.934\\
%     Mean(30\% of RGB) & 0.799 & 0.852\\
%     Mean(50\% of RGB) & 0.767 & 0.819\\
%     Mean(70\% of RGB) & 0.728 & 0.775\\
%     Mean(30\% of both) & 0.746 & 0.811\\
%     Mean(50\% of both) & 0.705 & 0.754\\
%     Mean(70\% of both) & 0.663 & 0.715\\
%   \bottomrule
%   \end{tabular}
  
%   }
%   \label{tab:more_modality_missing_situations}
%   \end{center}
%   \vspace{10pt}
% \end{table}

\vspace{-5pt}
\subsection{In-Depth Analysis}
\vspace{-2pt}

\noindent$\textbf{Comparisons with RGB-only, 3D-only methods.}$ Although our \textbf{\textit{MIIAD Bench}} does not follow a ``Multi-modal Training, Few-modal Inference'' setting, we rigorously conducted comparative experiments with RGB-only IAD models (using only 2D information) and 3D-only IAD models (relying solely on depth information), as shown in Tab.~\ref{tab:compare_2D_3D}. In this comprehensive experiment, all models were trained on modality-complete data. When comparing with RGB-only models, the 2D RGB modality remains complete while the 3D point cloud data is entirely missing. Conversely, when evaluating against 3D-only models, the 3D point cloud modality stays complete with completely missing 2D RGB data. This presents a significant challenge for \textbf{RADAR}, yet the table shows our model achieves the best AUPRO ($0.944$) and P-AUROC ($0.989$) in RGB-only scenarios. In 3D-only conditions, our model ranks second only to the specialized CMDIAD~\cite{sui2024incompletemultimodalindustrialanomaly} model designed for this setting, achieving optimal P-AUROC ($0.984$) and suboptimal AUPRO ($0.932$). These results demonstrate our model's capability to handle extreme cases of complete modality absence.

\begin{figure}[!t]
\includegraphics[width=0.5\textwidth]{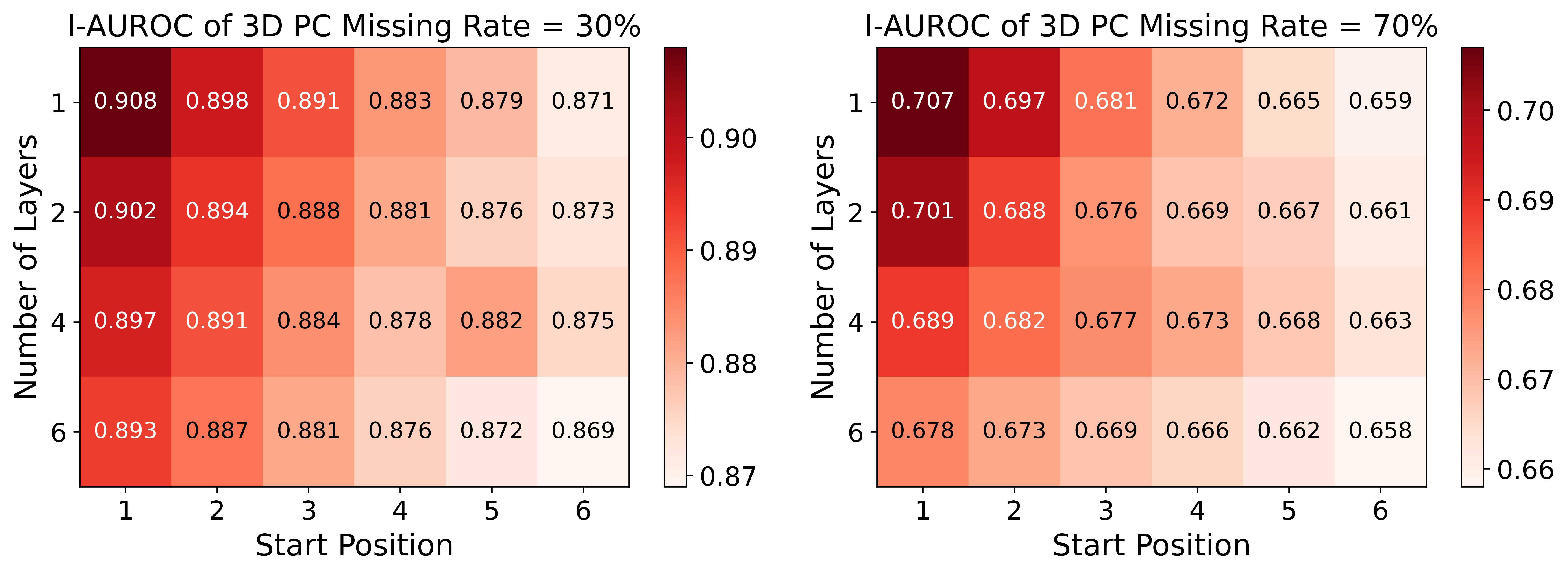}
\vspace{-20pt}
\centering\caption{Effect of the position of instruction layers.} 
\label{fig:Effect of the position of instruction layers}
\vspace{-18pt}
\end{figure}

\vspace{3pt}
\noindent$\textbf{Robustness of modality-incomplete in more situations.}$ In \\\textbf{\textit{MIIAD Bench}}, the modality missing conditions during training and testing are identical, so we further investigated scenarios where these conditions differ. As shown in Tab.~\ref{tab:radar_in_more_situations}, for a $30\%$ modality missing rate, this means both 2D RGB and 3D point cloud modalities have $15\%$ missing during either training or testing (totaling $30\%$), while the corresponding training or testing data remains modality-complete. The same applies to the $70\%$ missing rate. The table shows our model still outperforms other MIAD models. Taking MVTec-3D AD Split as an example, when test data has $30\%$ modality missing rate, our model improves I-AUROC and AUPRO by $2.4$ points ($2.7\%$) and $1.0$ points ($1.1\%$) respectively compared to the M3DM. With $30\%$ missing rate in training data, these improvements become $2.4$ points ($2.8\%$) and $1.6$ points ($1.8\%$). Similar performance is observed at $70\%$ missing rate, further demonstrating our model's robustness.

\vspace{3pt}
\noindent$\textbf{Computational costs.}$ We reduce computational costs by freezing the multimodal Transformer parameters and only training downstream task-related parameters. Experiments show this reduces parameters by $98.66\%$ versus the baseline. Our parameter efficiency reflects the real-time capability of our approach for practical MIIAD tasks. See Appendix Sec.~3 for details. It also explains why current MLLMs struggle with MIIAD tasks: they cannot effectively process 3D point clouds and suffer from excessive computational costs.

\begin{figure}[!t]
\includegraphics[width=0.4\textwidth]{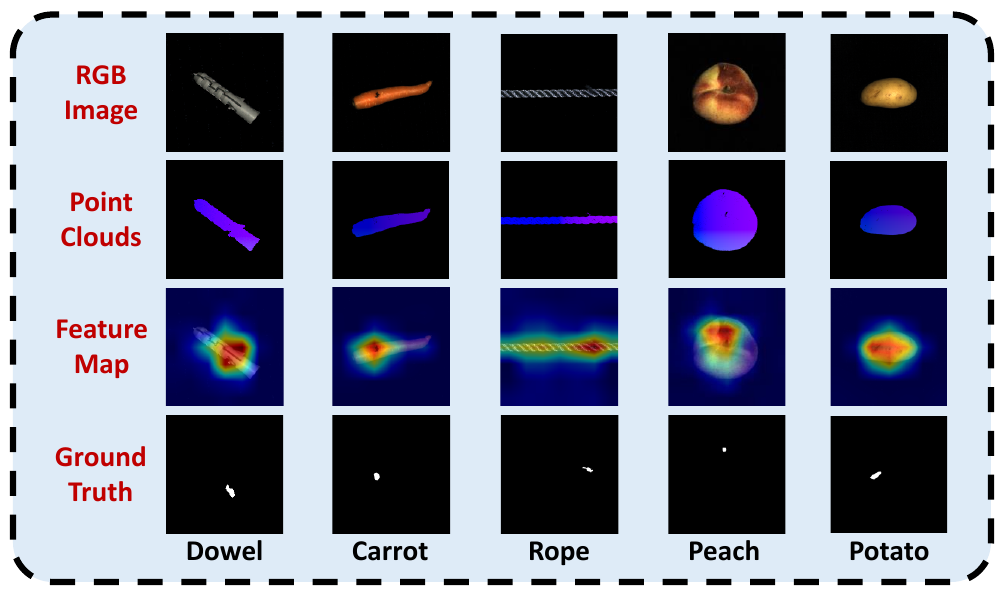}
\vspace{-10pt}
\centering\caption{Visualization results of IAD.} 
\vspace{-10pt}
\label{fig:visualization_results}
\vspace{-3pt}
\end{figure}

\vspace{5pt}
\noindent$\textbf{Effect of the position of instruction layers.}$ To thoroughly analyze the effectiveness of modality-incomplete instruction insertion positions, we visualize the effects of inserting instructions at different Transformer layers under various modality-missing scenarios, as shown in Fig.~\ref{fig:Effect of the position of instruction layers}. The results demonstrate that the choice of instruction layers significantly impacts model performance. Inserting modality-incomplete instructions at earlier layers with fewer inserted layers yields better results. This likely occurs because earlier-layer features retain more modality-specific characteristics compared to deeper layers. With instructions, multimodal inputs begin fusion from the very first Transformer layer, with fusion intensity increasing at deeper layers. Consequently, modality-incomplete instructions prove more effective when applied to earlier layers. Recent studies~\cite{lee2023cvpr} have provided similar insights.

\vspace{5pt}
\noindent$\textbf{Visualization results of IAD.}$  Fig. \ref{fig:visualization_results} illustrates IAD visualization results of our model. It showcases our model's aptitude for accurately identifying anomalies across diverse categories.

% \noindent$\textbf{Efficiency on parameters.}$ Our proposed method involves freezing all parameters of the multimodal Transformer, focusing solely on training parameters pertinent to downstream tasks. Further discussion on the efficacy of these parameters will be provided in the appendix.

% \noindent\textbf{Shortcomings and future work.}  Although our \textbf{RADAR} model enhances the performance and robustness of multimodal Transformers in MIIAD tasks, it does not reconstruct the lost data. Moving forward, we plan to delve into more advanced techniques like prompt-tuning~\cite{lester-etal-2021-power} and prefix-tuning~\cite{li-liang-2021-prefix} to enhance the design of modality-incomplete instruction.
% \vspace{-5pt}
\section{Conclusion}
\label{sec:conclusion} We propose the first-kind-of work to comprehensively investigate the \textbf{\underline{M}}odality-\textbf{\underline{I}}ncomplete \textbf{\underline{I}}ndustrial \textbf{\underline{A}}nomaly \textbf{\underline{D}}etection (\textbf{MIIAD}) with corresponding benchmark \textbf{\textit{MIIAD Bench}}, and construct a corresponding novel \textbf{\underline{R}}obust mod\textbf{\underline{A}}lity-instructive fusing and \textbf{\underline{D}}etecting fr\textbf{\underline{A}}mewo\textbf{\underline{R}}k, abbreviated as \textbf{RADAR}. 
% We first use ViT to extract 2D RGB image features, while introducing MaskTransformer and Point-MAE Decoder to extract 3D point cloud features; Afterwards, the modality-incomplete instruction are introduced into a concise multimodal Transformer, and a HyperNetwork is used to guide adaptive parameter learning, both of which boost the model to robustly cope with different modality-incomplete; Finally, we leverages Mahalanobis distance matrix and OCSVM to complete prediction based on the stored feature in multiple repositories. 
% We further explored the impact of prompt design and the robustness of the model in the absence of modality.
Extensive experiments indicated that our model outperforms other methods on \textbf{\textit{MIIAD Bench}}. We believe that
the proposed MIIAD task and \textbf{RADAR} serve as a complement to existing literature and provide new insights to the industrial anomaly detection community.

\section{Acknowledgment}
\label{acknowledgment}

This work has been supported in part by the NSFC (No. 62402426, 62272411, 62441617), the  Key Research and Development Projects in Zhejiang Province (No. 2025C01128, 2024C01106, 2025C01030, 2025C02156), Ningbo Yongjiang Talent Introduction Programme (2023A-400-G), Fundamental Research Funds for the Central Universities (226-2025-00057), Zhejiang University Education Foundation Qizhen Scholar Foundation.
% \clearpage

\bibliographystyle{ACM-Reference-Format}
\bibliography{main}

\clearpage
\appendix

\section*{Appendix}

This is the Appendix for the paper ``Robust Modality-Incomplete Anomaly Detection: a Modality-Instructive Framework with Benchmark'', which provides supplementary materials, detailed proofs, and additional experimental results to support the findings presented in the main manuscript. Tab.~\ref{tab:abb} shows abbreviations and symbols used in the main paper.

\vspace{-0.2cm}

\begin{table}[h]
\begin{center}
\captionsetup{font={small,stretch=1.25}, labelfont={bf}}
\caption{Abbreviations and symbols used in the main paper.}
 \renewcommand{\arraystretch}{1.2}
\resizebox{0.5\textwidth}{!}
 {
  \begin{tabular}{c||c}
   \toprule[1.5pt]
   \textbf{Abbreviation / Symbol} & \textbf{Meaning}\\
   \hline
   \hline
    \underline{\emph{Abbreviation}} & \\
    \textbf{RADAR} & Robust modAlity-instructive fusing and Detecting frAmewoRk\\
    MIIAD & Modality-Incomplete Industrial Anomaly Detection\\
    AIF & Adaptive Instruction Fusion\\
    MII & Modality-Incomplete Instruction\\
    ALM & Adaptive Learning Module\\
    DPHM & Double-Pseudo Hybrid Module\\
    MLP & Multi-Layer Perceptron\\
    \hline
    \hline
    
    \underline{\emph{Symbol in Algorithm}} & \\
    $\MIIADDataset^C$  & Modality-Complete Subset Data\\
    $\tilde{\MIIADDataset}^{2D}$  & 3D Point Cloud Modality-Incomplete Subset Data\\
    $\tilde{\MIIADDataset}^{3D}$  & 2D RGB Image Modality-Incomplete Subset Data\\
    $I_{m}$ & Modality-Incomplete Instructions\\
    $I_{pc}$ & MII for 3D Point Cloud Modality-Missing\\
    $I_{rgb}$ & MII for 2D RGB Image Modality-Missing\\
    $f_{pc}$ & Features of 3D Point Clouds\\
    $f_{rgb}$ & Features of 2D RGB Images\\
    $BCE$ & Binary Cross-Entropy\\
    % $\oplus$  & Concatenation\\
    % $F_{instruct}$  & Modality-incomplete instruction function\\
    $OCSVM$  & One-Class Support Vector Machine\\
    $MDM$  & Mahalanobis Distance Matrix\\
   \toprule[1.5pt]
   
  \end{tabular}
  }
  \label{tab:abb}
\end{center}
\vspace{-0.5cm}
\end{table}

In this supplementary material we present:

\begin{itemize}

\item Section~\ref{sec:implementation_details} provides detailed information on the \textbf{RADAR} model construction and training processes not covered in the main paper.

\item Section~\ref{sec:experimental_setting} provides a detailed description of the experimental settings in the main paper, including comparisons with both MIAD methods and missing-modality based approaches.

\item Section~\ref{sec:efficency_on_parameters} analyzes the parameter efficiency of \textbf{RADAR}, demonstrating that the framework requires minimal computational overhead.

\item Section~\ref{sec:pseudocode} provides the algorithmic pseudocode for \textbf{RADAR}, abstracting the pipeline workflow of the framework.

\item Section~\ref{sec:comparison_MIAD} presents expanded comparative experiments between \textbf{RADAR} and MIAD methods.

\item Section~\ref{sec:comparison_incomplete} provides additional comparisons of \textbf{RADAR} with missing-modality based approaches.

\item Section~\ref{sec:quantitative_results} supplements qualitative experiments for P-AUROC and AUPRO metrics.

\item Section~\ref{sec:ablation} elivers more comprehensive ablation studies.

\item Section~\ref{sec:hyper_parameters} conducts systematic in-depth analysis of hyperparameter ablation studies and selection.

\item Section~\ref{sec:more_visualization_results} offers extended visualization results for industrial anomaly detection.

\end{itemize}

\section{Implementation Details}
\label{sec:implementation_details}

The backbone network of \textbf{RADAR} consists of the following pre-trained components: \textbf{i) For 3D point clouds}, we adopt a combination of MaskTransformer~\cite{yu2022pointbert} and Point-MAE Decoder~\cite{pang2022masked} architectures to extract features. MaskTransformer uses transformer-based masked point modeling to capture long-range contextual relationships, while Point-MAE employs a masked autoencoding paradigm that effectively learns geometric representations from incomplete point clouds, making them particularly effective for 3D feature extraction. \textbf{ii) For 2D RGB images}, we use a Vision Transformer (ViT), specifically the ViT-B/8 model pre-trained on ImageNet~\cite{deng2009cvpr} with DINO framework that demonstrates strong anomaly detection capabilities. Both 2D\&3D feature extraction backbones are pre-trained models with powerful feature extraction capabilities, requiring no additional retraining.

% \vspace{5pt}
During data loading, when modalities are incomplete: \textbf{i) For missing 3D point cloud data} (encapsulated in TIFF format), we generate a pseudo-input $\tilde{x}^{m_1}$ as a tensor filled with ones matching original data dimensions. \textbf{ii) For missing 2D RGB image data}, we create a pseudo-image $\tilde{x}^{m_2}$ with all pixel values set to $1$. We use RANSAC~\cite{fischler1981random} to estimate background planes in 3D point clouds, removing points within $0.005$ distance threshold. Following M3DM~\cite{wang2023multimodal}, we resize both positional tensors and 2D images to $224 \times 224$, and convert outputs from layers $\{3, 7, 11\}$ into $x$, $y$, $z$ coordinates to represent 3D features.

% 

% \vspace{5pt}
During training, we freeze all parameters of the multimodal Transformer backbone and only train downstream task MIIAD-related parameters. Learnable components in \textbf{RADAR} include: modality-incomplete instruction, HyperNetwork (implemented as $2$-layer MLP), MLPs ($MLP_{\text{rgb}}$ and $MLP_{\text{pc}}$) in Stage1-Instructed Feature Fusion, multimodal decoders in DPHM Branch 1 (2D decoder with $3$ transposed conv layers; 3D decoder with MLP), upsampling network in DPHM Branch 1, and OCSVM in Stage2. Most components are lightweight (e.g., implemented with $1-2$ MLP layers). Below, we will discuss the implementation details of several main learnable modules: \textbf{i)} Modality-Incomplete Instruction, with a default length $L_i$ set to $16$. The MSA layer index with the prepended instruction ranges from $0$ to $3$. The optimizer utilized is AdamW, with the learning rate set to $1 \times 10^{-2}$. \textbf{ii)} $MLP_{\text{rgb}}$ and $MLP_{\text{pc}}$: use AdamW as the optimizer, a learning rate of $1 \times 10^{-3}$, and a batch size of $16$. \textbf{iii)} OCSVM: is optimized using SGD, with a learning rate of $1 \times 10^{-4}$. Further details are available in the appendix.

The threshold $\sigma$ for generating pseudo-labels $y_p$ in the DPHM Branch 1 is set as $0.35$. we directly took 30\% of the data samples and analyzed the anomaly score distribution of normal samples. We obtained all anomaly scores $a_i$, then took the 95\% as our reference threshold $\sigma$ (where 95\% of normal samples' $a_i$ are below this value), ultimately obtaining a value of $0.35$.

\section{Experimental Setting}
\label{sec:experimental_setting}

Beyond the comparisons with MIAD methods and missing-modality based methods discussed in the main text, our \textbf{RADAR} model has been extensively evaluated against additional relevant approaches.

\subsection{MIAD Methods}
To demonstrate how our \textbf{RADAR} model improves existing advanced MIAD methods, we conducted comprehensive comparative experiments on \textbf{\textit{MIIAD Bench}} with Shape-Guided~\cite{pmlr-v202-chu23b}, M3DM~\cite{wang2023multimodal}, AST~\cite{Rudolph_2023_WACV}, BTF~\cite{horwitz2023back}, AnomalyGPT~\cite{gu2023anomalyagpt} which is based on MLLM~\cite{zhang2024hyperllava, zhang2022boostmis, li2023fine, 10121664, li2022fine, lin2025healthgpt, yuan2024videorefer, gao2025benchmarkingmultimodalcotreward}, Gemini Vision Pro (using specific prompts), and GPT-4 Vision (using specific prompts). Below we briefly introduce these models (excluding Gemini Vision Pro and GPT-4 Vision):

\vspace{2pt}
\textbf{Shape-Guided~\cite{pmlr-v202-chu23b}:} Integrates color and geometric modalities using two specialized expert models and dual memory banks to achieve state-of-the-art performance in 3D anomaly detection.

\vspace{2pt}
\textbf{M3DM~\cite{wang2023multimodal}:} Proposes a multimodal industrial anomaly detection method with hybrid fusion, utilizing unsupervised feature fusion and decision layer fusion to enhance detection and segmentation performance on 3D point clouds and RGB images.

\vspace{2pt}
\textbf{AST~\cite{Rudolph_2023_WACV}:} Proposes asymmetric student-teacher networks for anomaly detection, using a bijective normalizing flow as the teacher and a conventional feed-forward network as the student to enhance anomaly detection performance.

\vspace{2pt}
\textbf{BTF~\cite{horwitz2023back}:} Proposes a method that combines classical 3D features (FPFH) with color-based features (PatchCore) to achieve state-of-the-art performance in 3D anomaly detection, emphasizing the importance of rotation invariance and complementary benefits of 3D and color modalities.

\vspace{2pt}
\textbf{AnomalyGPT~\cite{gu2023anomalyagpt}:} Leverages large vision-language models (LVLMs) to detect and localize industrial anomalies by fine-tuning on simulated anomaly data and employing a lightweight decoder for pixel-level localization, eliminating the need for manual threshold adjustments.

While general foundation models like \textbf{Gemini Vision Pro} and \textbf{GPT-4 Vision} possess broad knowledge, their inability to process 3D point clouds and the contextual differences across industrial anomalies limit their effectiveness for diverse industrial inspection needs. Following~\cite{xu2024custimizing}, we convert 3D point clouds into depth maps through planar projection, transforming them into standard 2D image formats readable by foundation models. We then design specific prompts including: \textit{Task instruction} (Clear task information prompts the generic foundation model for effective anomaly detection), \textit{Class context} (Explicit class context information, represented by a class token [CLS], enhances the model's recognition of the target domain), and \textit{Normality criteria} (Language-form rules define explicit normal standards and describe abnormal objects and patterns based on human expertise). The detailed prompt template is as follows:

\begin{tcolorbox}[
    colback=gray!10!white,
    colframe=gray!50!black,
    title=Prompt template for Gemini Vision Pro and GPT-4 Vision,
    fonttitle=\bfseries,
    boxrule=0.5mm,
    arc=2mm,
    width=\columnwidth,
    breakable,
    before skip=2mm,
    after skip=2mm,
    left=3pt,
    right=3pt,
    top=3pt,
    bottom=3pt
]
{\ttfamily\small
Variables: \\
!<INPUT 0>! -- Input type\\
!<INPUT 1>! -- Input Description Prompt\\
!<INPUT 2>! -- Normal Image\\
!<INPUT 3>! -- Anomaly Image\\

<commentblockmarker></commentblockmarker>\\
The first image given about the !<INPUT 0>! is normal. Please determine whether the second image given about the !<INPUT 0>! contains anomalies or defects. If yes, give a specific reason. Normally, the image given should depict clear and identifiable !<INPUT 0>!. It may have defects such as broken parts, and contaminations.\\

Your answer should contain 2 parts:\\

"reasoning": a string, that describes what you think about the given image;
"correctness": an integer, 1 if there is no anomaly in the image, 0 if it is, no partial credit;\\

Output example: \{"reasoning": a string, "correctness": integer, 1 or 0\}\\

Your response must be a valid JSON string starting with and should contain nothing else because we will directly execute it in Python. No indent or "n" in json structure formatting is needed. No ``` or ```json is needed.
}
\end{tcolorbox}

\subsection{Missing-modality Based Methods}

To demonstrate how our \textbf{RADAR} model outperforms missing-modality based methods, we conducted comparative experiments on \textbf{\textit{MIIAD Bench}} using M3DM as baseline, evaluating against CMDIAD~\cite{sui2024incompletemultimodalindustrialanomaly}, ShaSpec~\cite{wang2023multi}, VLMo~\cite{bao2022vlmo}, and GMC~\cite{poklukar2022gmc}. Below we briefly introduce these models:

\vspace{2pt}
\textbf{CMDIAD~\cite{sui2024incompletemultimodalindustrialanomaly}:} Proposes a Cross-Modal Distillation framework for Industrial Anomaly Detection, enabling multimodal training with few-modal inference to handle incomplete modalities in real-world scenarios.

\vspace{2pt}
\textbf{ShaSpec~\cite{wang2023multi}:} Devises a "Shared Specific Feature Modeling" method focused on learning shared and specific features across multimodal data to address missing modality representations.

\vspace{2pt}
\textbf{VLMo~\cite{bao2022vlmo}:} Introduces a hybrid multimodal expert model that leverages both the Dual Encoder for image retrieval tasks and the Fusion Encoder for multimodal encoding.

\vspace{2pt}
\textbf{GMC~\cite{poklukar2022gmc}:} Devises a geometric multimodal contrastive learning approach to tackle the challenge of inherent heterogeneity in multimodal learning.

% \vspace{2pt}
% \textbf{SMIL~\cite{ma2021aaai}:} Incorporates Bayesian meta-learning to explore the potential feature space, aiming to align the features of an incomplete mode with those of a complete mode.

\begin{table}[b]
    \vspace{-15pt}
    \begin{center}
    \captionsetup{font={small,stretch=1.25}, labelfont={bf}}
    \caption{\textbf{Comparison of parameter efficiency.}}
    \vspace{-10pt}
     \renewcommand{\arraystretch}{1.2}
     \resizebox{0.4\textwidth}{!}{
      \begin{tabular}{c|c c}
       \toprule[1.5pt]
       
        \textbf{Methods} &
        \textbf{\makecell{\# Trained\\params}} &
        \textbf{\makecell{Relative\\proportion}} \\
        \hline
        
        M3DM (Baseline)~\cite{wang2023multimodal} & 510,603K & 100\% \\
        
        \textbf{\textbf{RADAR} (Ours)} & 6,842K & 1.34\% \\
        \toprule[1.5pt]
      \end{tabular}
      }
      \label{tab:parameter_efficiency}
    \end{center}
    % \vspace{-20pt}
\end{table}

\section{Efficiency on Parameters}
\label{sec:efficency_on_parameters}

We have integrated HyperNetwork~\cite{ref:hypernetworks} to dynamically cater to tasks with fewer parameters, aiming for enhanced performance~\cite{karimi-mahabadi-etal-2021-parameter}. During the training of our modality-incomplete instruction, we leveraged prompt learning by freezing all parameters of the simple Multimodal Transformer, retaining only the MLP and fully connected layer for downstream task processing. Consequently, the number of parameters that our model \textbf{RADAR} requires to train is significantly reduced. As illustrated in the Tab. \ref{tab:parameter_efficiency}, in contrast to the $510$M parameter count in the Unsupervised Feature Fusion (UFF) module of the Baseline model M3DM~\cite{wang2023multimodal}, our model only necessitates training $6,842$K parameter counts, representing merely \textbf{$1.34\%$}. Nonetheless, our model's performance significantly exceeds that of the Baseline.

\begin{algorithm}[t]
\begin{flushleft}
\caption{Robust modality-instructive fusing and detecting framework (RADAR)}
\label{alg:pseudo_code}
\textbf{Input}: 
Multimodal dataset $\tilde{\MIIADDataset} = \{\MIIADDataset^C, \MIIADDataset^{2D}, \MIIADDataset^{3D}\}$ \\
\textbf{Output}: Anomaly score $\tau$, segmentation map $Seg_m$ \\

\textbf{Stage I: Adaptive Instruction Fusion} \\
\For{each sample $(d_{pc}, d_{rgb}) \in \tilde{\MIIADDataset}$}{
    \eIf{3D point cloud $d_{pc}$ missing}{
        $\tilde{x}^{m1} \leftarrow \text{ones\_tensor}$ \tcp{Pseudo 3D input}
        $f_{pc} \leftarrow \text{MaskTransformer}(\tilde{x}^{m1})$ \\
    }{
        $f_{pc} \leftarrow \text{Point-MAE}(\text{FPS}(d_{pc}))$ \tcp{Feature extraction}
    }
    \eIf{2D RGB $d_{rgb}$ missing}{
        $\tilde{x}^{m2} \leftarrow \text{ones\_image}$ \tcp{Pseudo 2D input}
        $f_{rgb} \leftarrow \text{ViT}(\tilde{x}^{m2})$ \\
    }{
        $f_{rgb} \leftarrow \text{ViT}(d_{rgb})$ \tcp{Feature extraction}
    }
    
    \tcp{Modality-Incomplete instruction}
    \uIf{both modalities complete}{$I_m \leftarrow I_{com}$}
    \uElseIf{only 3D missing}{$I_m \leftarrow I_{rgb}$}
    \uElseIf{only 2D missing}{$I_m \leftarrow I_{pc}$}
    
    \tcp{Adaptive parameter learning}
    $\{W^{(n)}, B^{(n)}\} \leftarrow \text{HyperNetwork}(f_{pc}, f_{rgb})$ \\
    $\hat{G} \leftarrow \text{Transformer}(\text{concat}(I_m, f_{pc}, f_{rgb}))$ \\
}

\textbf{Stage II: Double-Pseudo Hybrid Detection} \\
\For{each fused feature $\hat{G}$}{
    \tcp{Branch 1: Global reconstruction}
    $(\hat{x}_{rgb}, \hat{x}_{pc}) \leftarrow \text{Decoder}(\hat{G})$ \\
    $\mathcal{L}_{recon} \leftarrow \lambda_{rgb}\|\hat{x}_{rgb}-x_{rgb}\|^2 + \lambda_{pc}\|\hat{x}_{pc}-x_{pc}\|^2$ \\
    
    \tcp{Branch 2: Local contrastive learning}
    $\{g_i\} \leftarrow \text{PatchPooling}(\hat{G})$ \\
    $\mathcal{L}_{contrast} \leftarrow \text{BCE}(\text{TopK}(R_{fs}, \{g_i\}))$ \\
    
    \tcp{Repository-based detection}
    $R_{pc} \leftarrow \text{Update}(f_{pc}), R_{rgb} \leftarrow \text{Update}(f_{rgb})$ \\
    $\tau \leftarrow \text{MDM}(R_{pc}, R_{rgb}, R_{fs})$ \\
    $Seg_m \leftarrow \text{OCSVM}(R_{pc}, R_{rgb}, R_{fs})$ \\
}

\Return{$\tau$, $Seg_m$}
\end{flushleft}
\end{algorithm}

\section{RADAR Pipeline Pseudocode}
\label{sec:pseudocode}

In this section, we present the algorithmic pipeline of our proposed \textbf{RADAR} for modality-incomplete industrial anomaly detection. The framework operates through two core stages: \textbf{i) Adaptive Instruction Fusion} that dynamically handles missing modalities via learnable instructions and HyperNetwork-based parameter adaptation. \textbf{ii) Double-Pseudo Hybrid Detection} that mitigates overfitting through joint reconstruction-contrastive optimization. 

Below, we provide the pseudocode (Algorithm~\ref{alg:pseudo_code}) that formalizes three key innovations: \textbf{i)} modality-incomplete instruction injection for conditional feature fusion, \textbf{ii)} adaptive dynamic layers for missing-modality robustness, and \textbf{iii)} repository-based anomaly scoring with MDM and OCSVM. The algorithm explicitly captures how \textbf{RADAR} processes both complete (2D+3D) and incomplete (2D-only/3D-only) samples while maintaining detection stability across all missing-rate scenarios defined in \textbf{\textit{MIIAD Bench}}.

\vspace{-3pt}
\section{Comprehensive Comparison with MIAD Methods}
\label{sec:comparison_MIAD}

The comparative experimental results between our \textbf{RADAR} model and the additional MIAD methods mentioned in Section~\ref{sec:experimental_setting} are presented in Tab.~\ref{tab:miiad_bench_with_MIAD_I-AUROC}, Tab.~\ref{tab:miiad_bench_with_MIAD_P-AUROC}, and Tab.~\ref{tab:miiad_bench_with_MIAD_AUPRO}. In addition to expanded comparisons with more MIAD methods, we have supplemented the P-AUROC results (Tab.~\ref{tab:miiad_bench_with_MIAD_P-AUROC}) missing from the main text.

These more comprehensive experimental results further validate our key claims in the main text: that MIIAD represents a novel, challenging, and practical task, and that our proposed \textbf{RADAR} framework effectively enhances the performance of MIAD methods.

\vspace{-3pt}
\section{Comprehensive Comparison with Missing\\-modality Based Methods}
\label{sec:comparison_incomplete}

The extensive comparative experimental results between our \textbf{RADAR} model and the additional missing-modality based methods mentioned in Section~2 are systematically presented in Tab.~\ref{tab:miiad_bench_with_incomplete_I-AUROC}, Tab.~\ref{tab:miiad_bench_with_incomplete_P-AUROC}, and Tab.~\ref{tab:miiad_bench_with_incomplete_AUPRO}, comprehensively covering all key evaluation metrics including I-AUROC, P-AUROC, and AUPRO.

These more extensive experimental results further validate our claims in the main text: current advanced missing-modality based methods still struggle with MIIAD challenges, while our proposed \textbf{RADAR} framework significantly outperforms these approaches.

\vspace{-3pt}
\section{Quantitative Results of P-AUROC and AUPRO}
\label{sec:quantitative_results}

\begin{figure*}[!t]
\includegraphics[width=0.98\textwidth]{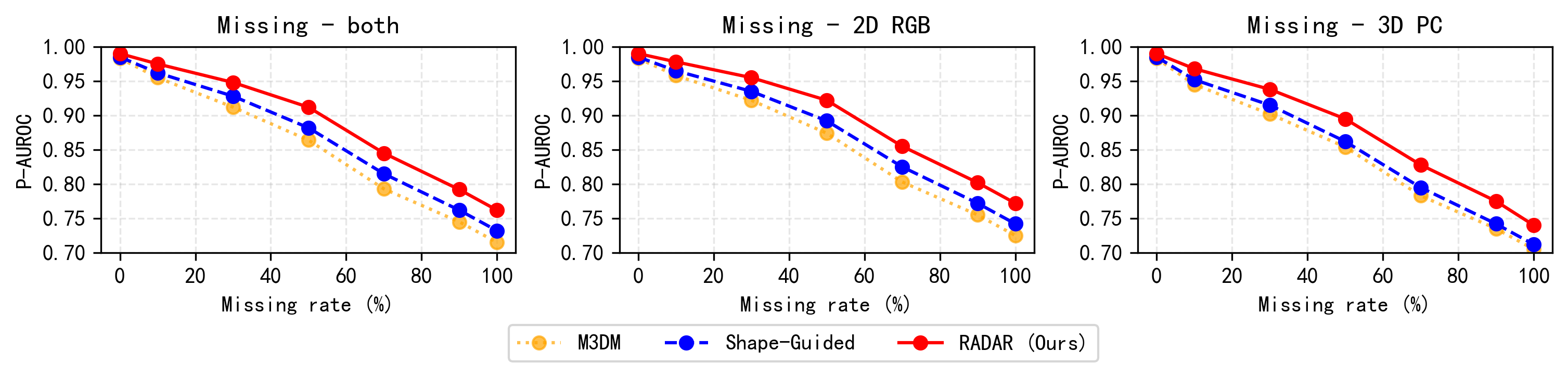}
\vspace{-15pt}
\centering\caption{Quantitative results of P-AUROC on the \textit{MIIAD Bench} - MVTEC-3D AD with different missing rates under different missing-modality scenarios. Each data point in the figure represents that training and testing are with the same $\eta\%$ missing rate.} 
\label{fig:performance_under_different_modal_missing_rates_P-AUROC}
% \vspace{-10pt}
\end{figure*}

\begin{figure*}[!t]
\includegraphics[width=0.98\textwidth]{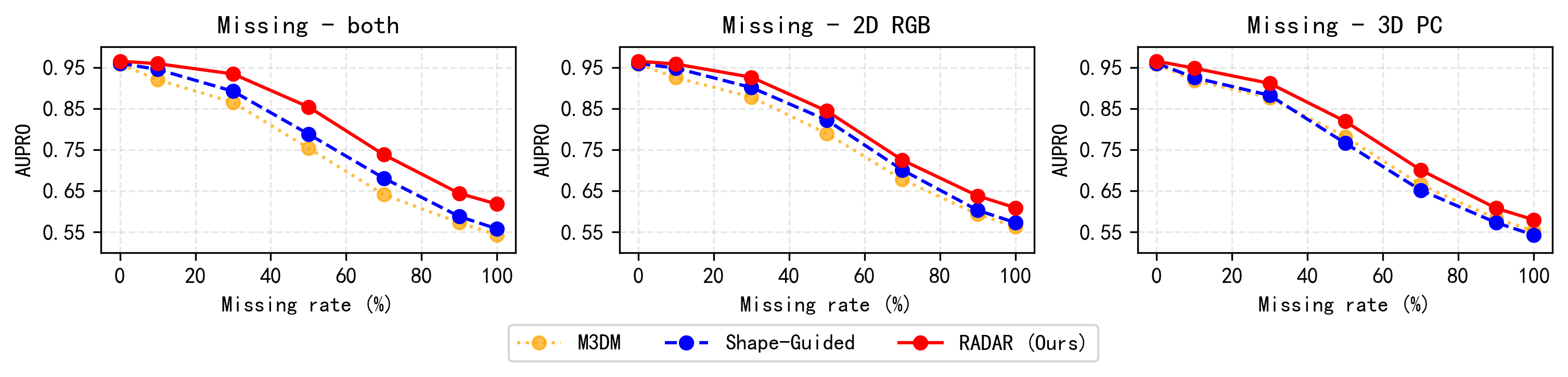}
\vspace{-15pt}
\centering\caption{Quantitative results of AUPRO on the \textit{MIIAD Bench} - MVTEC-3D AD with different missing rates under different missing-modality scenarios. Each data point in the figure represents that training and testing are with the same $\eta\%$ missing rate.} 
\label{fig:performance_under_different_modal_missing_rates_AUPRO}
\vspace{-5pt}
\end{figure*}

Furthermore, Fig.~\ref{fig:performance_under_different_modal_missing_rates_P-AUROC} and Fig.~\ref{fig:performance_under_different_modal_missing_rates_AUPRO} qualitatively demonstrate the performance trends of different models' P-AUROC and AUPRO metrics under various settings as modality missing rates increase. The results reveal phenomena consistent with those observed for I-AUROC in the main text: while current MIAD methods show decreasing P-AUROC and AUPRO scores with increasing modality missing rates, our \textbf{RADAR} method still achieves significant performance improvements over the baseline.

\begin{table*}[t]
  \begin{center}
  \captionsetup{font={small,stretch=1.25}, labelfont={bf}}
  \caption{I-AUROC score in performance comparison of different MIAD methods on \textbf{\textit{MIIAD Bench}}. \textcolor{deepred}{\textbf{Red bold values}} indicate the best experimental performance, while \underline{underlined values} denote the second-best results.}
  \vspace{-10pt}
   \renewcommand{\arraystretch}{1.2}
   \resizebox{1\textwidth}{!}{
    \begin{tabular}{c|c||c c||c c||c c c c c c c c}
     \toprule[1.5pt]
     
      \multirow{2}{*}[-0.5ex]{\makecell[c]{\textbf{Missing}\\\textbf{Rate}}} &
      \multirow{2}{*}[-0.5ex]{\makecell[c]{\textbf{\textit{MIIAD Bench}}\\\textbf{Split}}}  & 
      \multicolumn{2}{c||}{\textbf{Data in \# Train}} & 
      \multicolumn{2}{c||}{\textbf{Data in \# Test}} & 
      \multicolumn{8}{c}{\textbf{I-AUROC}} \\
      \cline{3-14}
      & & \textbf{3D PC} & \textbf{2D RGB} & \textbf{3D PC} & \textbf{2D RGB} & 
      \textbf{Gemini-V} & \textbf{GPT-4V} & \textbf{BTF}~\cite{horwitz2023back} & \textbf{AST}~\cite{Rudolph_2023_WACV} & \textbf{AnomalyGPT}~\cite{gu2023anomalyagpt} & \textbf{M3DM}~\cite{wang2023multimodal} & \textbf{Shape-Guided}~\cite{pmlr-v202-chu23b} & \makecell[c]{\textbf{RADAR}\\\textbf{(Ours)}} \\\hline\hline

      \multirow{2}{*}{0\%}
      & \makecell[c]{MVTec-3D AD} & \multicolumn{2}{c||}{100\%} & \multicolumn{2}{c||}{100\%} & 0.574 & 0.753 & 0.865 & 0.937 & 0.922 & 0.945 & \textcolor{deepred}{\textbf{0.947}} & \textcolor{deepred}{\textbf{0.947}} \\\cline{2-14}
      & \makecell[c]{Eyecandies} & \multicolumn{2}{c||}{100\%} & \multicolumn{2}{c||}{100\%} & 0.503 & 0.676 & 0.816 & 0.879 & 0.857 & \underline{0.897} & 0.891 & \textcolor{deepred}{\textbf{0.901}} \\
      \hline

      \multirow{6}{*}{30\%} 
      & \multirow{3}{*}{\makecell[c]{MVTec-\\3D AD}}
      & 100\% & 70\% & 100\% & 70\% & 0.521 & 0.682 & 0.813 & 0.869 & 0.841 & 0.892 & \textcolor{deepred}{\textbf{0.910}} & \underline{0.908} \\
      & & 70\% & 100\% & 70\% & 100\% & 0.506 & 0.693 & 0.795 & \underline{0.901} & 0.828 & 0.883 & 0.869 & \textcolor{deepred}{\textbf{0.903}} \\
      & & 85\% & 85\% & 85\% & 85\% & 0.515 & 0.684 & 0.797 & 0.873 & 0.826 & \underline{0.886} & 0.883 & \textcolor{deepred}{\textbf{0.907}} \\\cline{2-14}
      & \multirow{3}{*}{\makecell[c]{Eye-\\candies}}
      & 100\% & 70\% & 100\% & 70\% & 0.462 & 0.624 & 0.749 & 0.826 & 0.784 & 0.844 & \underline{0.864} & \textcolor{deepred}{\textbf{0.869}} \\
      & & 70\% & 100\% & 70\% & 100\% & 0.457 & 0.615 & 0.732 & \textcolor{deepred}{\textbf{0.858}} & 0.778 & 0.843 & 0.837 & \textcolor{deepred}{\textbf{0.858}} \\
      & & 85\% & 85\% & 85\% & 85\% & 0.464 & 0.620 & 0.745 & 0.832 & 0.783 & 0.843 & \underline{0.850} & \textcolor{deepred}{\textbf{0.867}} \\
      \hline

      \multirow{6}{*}{50\%} 
      & \multirow{3}{*}{\makecell[c]{MVTec-\\3D AD}}
      & 100\% & 50\% & 100\% & 50\% & 0.446 & 0.591 & 0.736 & 0.772 & 0.732 & 0.796 & \textcolor{deepred}{\textbf{0.811}} & \underline{0.810} \\
      & & 50\% & 100\% & 50\% & 100\% & 0.459 & 0.584 & 0.718 & \underline{0.807} & 0.724 & 0.789 & 0.773 & \textcolor{deepred}{\textbf{0.821}} \\
      & & 75\% & 75\% & 75\% & 75\% & 0.450 & 0.587 & 0.726 & 0.782 & 0.718 & 0.791 & \underline{0.799} & \textcolor{deepred}{\textbf{0.817}} \\\cline{2-14}
      & \multirow{3}{*}{\makecell[c]{Eye-\\candies}}
      & 100\% & 50\% & 100\% & 50\% & 0.414 & 0.565 & 0.685 & 0.742 & 0.703 & 0.765 & \underline{0.786} & \textcolor{deepred}{\textbf{0.795}} \\
      & & 50\% & 100\% & 50\% & 100\% & 0.423 & 0.549 & 0.679 & \underline{0.764} & 0.695 & 0.757 & 0.756 & \textcolor{deepred}{\textbf{0.780}} \\
      & & 75\% & 75\% & 75\% & 75\% & 0.419 & 0.554 & 0.674 & 0.754 & 0.701 & 0.759 & \underline{0.765} & \textcolor{deepred}{\textbf{0.787}} \\
      \hline

      \multirow{6}{*}{70\%} 
      & \multirow{3}{*}{\makecell[c]{MVTec-\\3D AD}}
      & 100\% & 30\% & 100\% & 30\% & 0.316 & 0.476 & 0.606 & 0.653 & 0.616 & 0.675 & \underline{0.685} & \textcolor{deepred}{\textbf{0.707}} \\
      & & 30\% & 100\% & 30\% & 100\% & 0.332 & 0.488 & 0.588 & \underline{0.671} & 0.608 & 0.662 & 0.649 & \textcolor{deepred}{\textbf{0.703}} \\
      & & 65\% & 65\% & 65\% & 65\% & 0.324 & 0.484 & 0.594 & 0.660 & 0.615 & 0.668 & \underline{0.672} & \textcolor{deepred}{\textbf{0.706}} \\\cline{2-14}
      & \multirow{3}{*}{\makecell[c]{Eye-\\candies}}
      & 100\% & 30\% & 100\% & 30\% & 0.284 & 0.437 & 0.563 & 0.601 & 0.571 & 0.629 & \underline{0.635} & \textcolor{deepred}{\textbf{0.662}} \\
      & & 30\% & 100\% & 30\% & 100\% & 0.271 & 0.429 & 0.552 & \underline{0.624} & 0.565 & 0.623 & 0.611 & \textcolor{deepred}{\textbf{0.654}} \\
      & & 65\% & 65\% & 65\% & 65\% & 0.279 & 0.433 & 0.559 & 0.614 & 0.570 & 0.624 & \underline{0.626} & \textcolor{deepred}{\textbf{0.658}} \\
      
      \toprule[1.5pt]
    \end{tabular}
    }
    \label{tab:miiad_bench_with_MIAD_I-AUROC}
  \end{center}
  % \vspace{-15pt}
\end{table*}

\begin{table*}[t]
  \begin{center}
  \captionsetup{font={small,stretch=1.25}, labelfont={bf}}
  \caption{P-AUROC score in performance comparison of different MIAD methods on \textbf{\textit{MIIAD Bench}}.}
  \vspace{-10pt}
   \renewcommand{\arraystretch}{1.2}
   \resizebox{1\textwidth}{!}{
    \begin{tabular}{c|c||c c||c c||c c c c c c c c}
     \toprule[1.5pt]
     
      \multirow{2}{*}[-0.5ex]{\makecell[c]{\textbf{Missing}\\\textbf{Rate}}} &
      \multirow{2}{*}[-0.5ex]{\makecell[c]{\textbf{\textit{MIIAD Bench}}\\\textbf{Split}}}  & 
      \multicolumn{2}{c||}{\textbf{Data in \# Train}} & 
      \multicolumn{2}{c||}{\textbf{Data in \# Test}} & 
      \multicolumn{8}{c}{\textbf{P-AUROC}} \\
      \cline{3-14}
      & & \textbf{3D PC} & \textbf{2D RGB} & \textbf{3D PC} & \textbf{2D RGB} & 
      \textbf{Gemini-V} & \textbf{GPT-4V} & \textbf{BTF}~\cite{horwitz2023back} & \textbf{AST}~\cite{Rudolph_2023_WACV} & \textbf{AnomalyGPT}~\cite{gu2023anomalyagpt} & \textbf{M3DM}~\cite{wang2023multimodal} & \textbf{Shape-Guided}~\cite{pmlr-v202-chu23b} & \makecell[c]{\textbf{RADAR}\\\textbf{(Ours)}} \\\hline\hline
    
      \multirow{2}{*}{0\%}
      & \makecell[c]{MVTec-3D AD} & \multicolumn{2}{c||}{100\%} & \multicolumn{2}{c||}{100\%} & 0.606 & 0.787 & \textcolor{deepred}{\textbf{0.992}} & 0.976 & 0.927 & \textcolor{deepred}{\textbf{0.992}} & 0.985 & \textcolor{deepred}{\textbf{0.992}} \\\cline{2-14}
      & \makecell[c]{Eyecandies} & \multicolumn{2}{c||}{100\%} & \multicolumn{2}{c||}{100\%} & 0.575 & 0.754 & \textcolor{deepred}{\textbf{0.981}} & 0.958 & 0.906 & 0.977 & 0.979 & \textcolor{deepred}{\textbf{0.981}} \\
      \hline
    
      \multirow{6}{*}{30\%} 
      & \multirow{3}{*}{\makecell[c]{MVTec-\\3D AD}}
      & 100\% & 70\% & 100\% & 70\% & 0.526 & 0.722 & \underline{0.945} & 0.916 & 0.857 & \underline{0.945} & 0.930 & \textcolor{deepred}{\textbf{0.955}} \\
      & & 70\% & 100\% & 70\% & 100\% & 0.518 & 0.715 & \underline{0.941} & 0.909 & 0.849 & 0.938 & 0.923 & \textcolor{deepred}{\textbf{0.948}} \\
      & & 85\% & 85\% & 85\% & 85\% & 0.532 & 0.728 & 0.942 & 0.913 & 0.861 & \underline{0.947} & 0.928 & \textcolor{deepred}{\textbf{0.953}} \\\cline{2-14}
      & \multirow{3}{*}{\makecell[c]{Eye-\\candies}}
      & 100\% & 70\% & 100\% & 70\% & 0.495 & 0.694 & \underline{0.933} & 0.898 & 0.836 & 0.928 & 0.919 & \textcolor{deepred}{\textbf{0.943}} \\
      & & 70\% & 100\% & 70\% & 100\% & 0.487 & 0.687 & \underline{0.926} & 0.891 & 0.828 & 0.921 & 0.912 & \textcolor{deepred}{\textbf{0.936}} \\
      & & 85\% & 85\% & 85\% & 85\% & 0.501 & 0.699 & \underline{0.930} & 0.895 & 0.839 & 0.925 & 0.917 & \textcolor{deepred}{\textbf{0.941}} \\
      \hline
    
      \multirow{6}{*}{50\%} 
      & \multirow{3}{*}{\makecell[c]{MVTec-\\3D AD}}
      & 100\% & 50\% & 100\% & 50\% & 0.441 & 0.632 & 0.865 & 0.826 & 0.767 & \underline{0.868} & 0.850 & \textcolor{deepred}{\textbf{0.888}} \\
      & & 50\% & 100\% & 50\% & 100\% & 0.433 & 0.625 & \underline{0.863} & 0.818 & 0.759 & 0.857 & 0.842 & \textcolor{deepred}{\textbf{0.880}} \\
      & & 75\% & 75\% & 75\% & 75\% & 0.447 & 0.638 & 0.860 & 0.823 & 0.771 & \underline{0.861} & 0.848 & \textcolor{deepred}{\textbf{0.885}} \\\cline{2-14}
      & \multirow{3}{*}{\makecell[c]{Eye-\\candies}}
      & 100\% & 50\% & 100\% & 50\% & 0.412 & 0.604 & \underline{0.853} & 0.808 & 0.746 & 0.847 & 0.839 & \textcolor{deepred}{\textbf{0.873}} \\
      & & 50\% & 100\% & 50\% & 100\% & 0.404 & 0.597 & \underline{0.845} & 0.800 & 0.738 & 0.839 & 0.831 & \textcolor{deepred}{\textbf{0.865}} \\
      & & 75\% & 75\% & 75\% & 75\% & 0.418 & 0.610 & \underline{0.849} & 0.805 & 0.751 & 0.843 & 0.836 & \textcolor{deepred}{\textbf{0.870}} \\
      \hline
    
      \multirow{6}{*}{70\%} 
      & \multirow{3}{*}{\makecell[c]{MVTec-\\3D AD}}
      & 100\% & 30\% & 100\% & 30\% & 0.351 & 0.532 & 0.744 & 0.716 & 0.657 & \underline{0.755} & 0.740 & \textcolor{deepred}{\textbf{0.785}} \\
      & & 30\% & 100\% & 30\% & 100\% & 0.343 & 0.525 & \underline{0.747} & 0.708 & 0.649 & 0.741 & 0.732 & \textcolor{deepred}{\textbf{0.777}} \\
      & & 65\% & 65\% & 65\% & 65\% & 0.357 & 0.538 & 0.751 & 0.713 & 0.661 & \underline{0.755} & 0.738 & \textcolor{deepred}{\textbf{0.782}} \\\cline{2-14}
      & \multirow{3}{*}{\makecell[c]{Eye-\\candies}}
      & 100\% & 30\% & 100\% & 30\% & 0.327 & 0.504 & \underline{0.743} & 0.698 & 0.636 & 0.737 & 0.729 & \textcolor{deepred}{\textbf{0.763}} \\
      & & 30\% & 100\% & 30\% & 100\% & 0.319 & 0.497 & \underline{0.735} & 0.690 & 0.628 & 0.729 & 0.721 & \textcolor{deepred}{\textbf{0.755}} \\
      & & 65\% & 65\% & 65\% & 65\% & 0.333 & 0.510 & \underline{0.739} & 0.695 & 0.641 & 0.733 & 0.726 & \textcolor{deepred}{\textbf{0.760}} \\
      
      \toprule[1.5pt]
    \end{tabular}
    }
    \label{tab:miiad_bench_with_MIAD_P-AUROC}
  \end{center}
  % \vspace{-15pt}
\end{table*}

\begin{table*}[t]
  \begin{center}
  \captionsetup{font={small,stretch=1.25}, labelfont={bf}}
  \caption{AUPRO score in performance comparison of different MIAD methods on \textbf{\textit{MIIAD Bench}}.}
  \vspace{-10pt}
   \renewcommand{\arraystretch}{1.2}
   \resizebox{1\textwidth}{!}{
    \begin{tabular}{c|c||c c||c c||c c c c c c}
     \toprule[1.5pt]
     
      \multirow{2}{*}[-0.5ex]{\makecell[c]{\textbf{Missing}\\\textbf{Rate}}} &
      \multirow{2}{*}[-0.5ex]{\makecell[c]{\textbf{\textit{MIIAD Bench}}\\\textbf{Split}}}  & 
      \multicolumn{2}{c||}{\textbf{Data in \# Train}} & 
      \multicolumn{2}{c||}{\textbf{Data in \# Test}} & 
      \multicolumn{6}{c}{\textbf{AUPRO}} \\
      \cline{3-12}
      & & \textbf{3D PC} & \textbf{2D RGB} & \textbf{3D PC} & \textbf{2D RGB} & 
      \textbf{Gemini-V} & \textbf{GPT-4V} & \textbf{BTF}~\cite{horwitz2023back} & \textbf{M3DM}~\cite{wang2023multimodal} & \textbf{Shape-Guided}~\cite{pmlr-v202-chu23b} & \makecell[c]{\textbf{RADAR (Ours)}}\\\hline\hline

      \multirow{2}{*}{0\%}
      & \makecell[c]{MVTec-3D AD} & \multicolumn{2}{c||}{100\%} & \multicolumn{2}{c||}{100\%} & 0.344 & 0.572 & 0.959 & 0.964 & \textcolor{deepred}{\textbf{0.976}} & \underline{0.967} \\\cline{2-12}
      & \makecell[c]{Eyecandies} & \multicolumn{2}{c||}{100\%} & \multicolumn{2}{c||}{100\%} & 0.297 & 0.514 & 0.866 & \underline{0.882} & 0.876 & \textcolor{deepred}{\textbf{0.885}} \\
      \hline

      \multirow{6}{*}{30\%} 
      & \multirow{3}{*}{\makecell[c]{MVTec-\\3D AD}}
      & 100\% & 70\% & 100\% & 70\% & 0.315 & 0.544 & 0.903 & 0.911 & \underline{0.924} & \textcolor{deepred}{\textbf{0.925}} \\
      & & 70\% & 100\% & 70\% & 100\% & 0.307 & 0.530 & \underline{0.937} & 0.932 & 0.918 & \textcolor{deepred}{\textbf{0.947}} \\
      & & 85\% & 85\% & 85\% & 85\% & 0.309 & 0.540 & 0.914 & 0.917 & \underline{0.923} & \textcolor{deepred}{\textbf{0.929}} \\\cline{2-12}
      & \multirow{3}{*}{\makecell[c]{Eye-\\candies}}
      & 100\% & 70\% & 100\% & 70\% & 0.264 & 0.453 & 0.813 & \underline{0.828} & 0.832 & \textcolor{deepred}{\textbf{0.847}} \\
      & & 70\% & 100\% & 70\% & 100\% & 0.248 & 0.458 & \underline{0.846} & 0.832 & 0.829 & \textcolor{deepred}{\textbf{0.851}} \\
      & & 85\% & 85\% & 85\% & 85\% & 0.252 & 0.454 & 0.821 & 0.830 & \underline{0.831} & \textcolor{deepred}{\textbf{0.854}} \\
      \hline

      \multirow{6}{*}{50\%} 
      & \multirow{3}{*}{\makecell[c]{MVTec-\\3D AD}}
      & 100\% & 50\% & 100\% & 50\% & 0.248 & 0.485 & 0.826 & 0.835 & \underline{0.848} & \textcolor{deepred}{\textbf{0.857}} \\
      & & 50\% & 100\% & 50\% & 100\% & 0.261 & 0.489 & \underline{0.857} & 0.849 & 0.838 & \textcolor{deepred}{\textbf{0.874}} \\
      & & 75\% & 75\% & 75\% & 75\% & 0.253 & 0.483 & 0.819 & \underline{0.833} & 0.831 & \textcolor{deepred}{\textbf{0.862}} \\\cline{2-12}
      & \multirow{3}{*}{\makecell[c]{Eye-\\candies}}
      & 100\% & 50\% & 100\% & 50\% & 0.202 & 0.387 & 0.731 & \underline{0.749} & 0.746 & \textcolor{deepred}{\textbf{0.772}} \\
      & & 50\% & 100\% & 50\% & 100\% & 0.206 & 0.375 & \underline{0.750} & 0.741 & 0.741 & \textcolor{deepred}{\textbf{0.769}} \\
      & & 75\% & 75\% & 75\% & 75\% & 0.212 & 0.380 & \underline{0.744} & 0.742 & 0.738 & \textcolor{deepred}{\textbf{0.771}} \\
      \hline

      \multirow{6}{*}{70\%} 
      & \multirow{3}{*}{\makecell[c]{MVTec-\\3D AD}}
      & 100\% & 30\% & 100\% & 30\% & 0.183 & 0.394 & 0.702 & 0.715 & \underline{0.729} & \textcolor{deepred}{\textbf{0.755}} \\
      & & 30\% & 100\% & 30\% & 100\% & 0.184 & 0.401 & \underline{0.718} & 0.709 & 0.696 & \textcolor{deepred}{\textbf{0.748}} \\
      & & 65\% & 65\% & 65\% & 65\% & 0.178 & 0.398 & 0.706 & 0.711 & \underline{0.714} & \textcolor{deepred}{\textbf{0.752}} \\\cline{2-12}
      & \multirow{3}{*}{\makecell[c]{Eye-\\candies}}
      & 100\% & 30\% & 100\% & 30\% & 0.154 & 0.334 & 0.596 & 0.613 & \underline{0.616} & \textcolor{deepred}{\textbf{0.647}} \\
      & & 30\% & 100\% & 30\% & 100\% & 0.162 & 0.328 & \underline{0.619} & 0.608 & 0.602 & \textcolor{deepred}{\textbf{0.649}} \\
      & & 65\% & 65\% & 65\% & 65\% & 0.159 & 0.331 & 0.601 & 0.610 & \underline{0.615} & \textcolor{deepred}{\textbf{0.649}} \\
      
      \toprule[1.5pt]
    \end{tabular}
    }
    \label{tab:miiad_bench_with_MIAD_AUPRO}
  \end{center}
  % \vspace{-15pt}
\end{table*}

\begin{table*}[t]
  \begin{center}
  \captionsetup{font={small,stretch=1.25}, labelfont={bf}}
  \caption{I-AUROC score in performance comparison of missing-modality based methods on \textbf{\textit{MIIAD Bench}}.}
  \vspace{-10pt}
   \renewcommand{\arraystretch}{1.2}
   \resizebox{1\textwidth}{!}{
    \begin{tabular}{c|c||c c||c c||c c c c c c}
     \toprule[1.5pt]
     
      \multirow{2}{*}[-0.5ex]{\makecell[c]{\textbf{Missing}\\\textbf{Rate}}} &
      \multirow{2}{*}[-0.5ex]{\makecell[c]{\textbf{\textit{MIIAD Bench}}\\\textbf{Split}}}  & 
      \multicolumn{2}{c||}{\textbf{Data in \# Train}} & 
      \multicolumn{2}{c||}{\textbf{Data in \# Test}} & 
      \multicolumn{6}{c}{\textbf{I-AUROC}} \\
      \cline{3-12}
      & & \textbf{3D PC} & \textbf{2D RGB} & \textbf{3D PC} & \textbf{2D RGB} & 
      \makecell[c]{\textbf{M3DM}\\\textbf{(Baseline)}}~\cite{wang2023multimodal} & \textbf{+GMC}~\cite{poklukar2022gmc} & \textbf{+VLMo}~\cite{bao2022vlmo} & \textbf{+ShaSpec}~\cite{wang2023multi} & \textbf{+CMDIAD}~\cite{sui2024incompletemultimodalindustrialanomaly} & \makecell[c]{\textbf{+RADAR}\\\textbf{(Ours)}}\\\hline\hline

      \multirow{2}{*}{0\%}
      & \makecell[c]{MVTec-3D AD} & \multicolumn{2}{c||}{100\%} & \multicolumn{2}{c||}{100\%} & 0.945 & 0.931 & 0.937 & 0.943 & \textcolor{deepred}{\textbf{0.948}} & \underline{0.947}  \\\cline{2-12}
      & \makecell[c]{Eyecandies} & \multicolumn{2}{c||}{100\%} & \multicolumn{2}{c||}{100\%} & 0.897 & 0.886 & 0.897 & 0.891 & \underline{0.898} & \textcolor{deepred}{\textbf{0.901}} \\
      \hline

      \multirow{6}{*}{30\%} 
      & \multirow{3}{*}{\makecell[c]{MVTec-\\3D AD}}
      & 100\% & 70\% & 100\% & 70\% & 0.892 & 0.878 & 0.895 & \underline{0.901} & 0.898 & \textcolor{deepred}{\textbf{0.908}} \\
      & & 70\% & 100\% & 70\% & 100\% & 0.883 & 0.868 & 0.888 & 0.887 & \underline{0.894} & \textcolor{deepred}{\textbf{0.903}} \\
      & & 85\% & 85\% & 85\% & 85\% & 0.886 & 0.872 & 0.892 & 0.893 & \underline{0.897} & \textcolor{deepred}{\textbf{0.907}} \\\cline{2-12}
      & \multirow{3}{*}{\makecell[c]{Eye-\\candies}}
      & 100\% & 70\% & 100\% & 70\% & 0.844 & 0.830 & \underline{0.848} & 0.847 & 0.842 & \textcolor{deepred}{\textbf{0.869}} \\
      & & 70\% & 100\% & 70\% & 100\% & 0.843 & 0.828 & 0.842 & \underline{0.851} & 0.845 & \textcolor{deepred}{\textbf{0.858}} \\
      & & 85\% & 85\% & 85\% & 85\% & 0.843 & 0.829 & 0.844 & \underline{0.852} & 0.843 & \textcolor{deepred}{\textbf{0.867}} \\
      \hline

      \multirow{6}{*}{50\%} 
      & \multirow{3}{*}{\makecell[c]{MVTec-\\3D AD}}
      & 100\% & 50\% & 100\% & 50\% & 0.796 & 0.776 & 0.792 & \underline{0.803} & 0.801 & \textcolor{deepred}{\textbf{0.810}} \\
      & & 50\% & 100\% & 50\% & 100\% & 0.789 & 0.770 & 0.787 & \underline{0.797} & 0.795 & \textcolor{deepred}{\textbf{0.821}} \\
      & & 75\% & 75\% & 75\% & 75\% & 0.791 & 0.771 & 0.789 & \underline{0.798} & 0.797 & \textcolor{deepred}{\textbf{0.817}} \\\cline{2-12}
      & \multirow{3}{*}{\makecell[c]{Eye-\\candies}}
      & 100\% & 50\% & 100\% & 50\% & 0.765 & 0.745 & 0.763 & \underline{0.772} & 0.770 & \textcolor{deepred}{\textbf{0.795}} \\
      & & 50\% & 100\% & 50\% & 100\% & 0.757 & 0.737 & 0.755 & \underline{0.764} & 0.762 & \textcolor{deepred}{\textbf{0.780}} \\
      & & 75\% & 75\% & 75\% & 75\% & 0.759 & 0.739 & 0.757 & \underline{0.766} & 0.764 & \textcolor{deepred}{\textbf{0.787}} \\
      \hline

      \multirow{6}{*}{70\%} 
      & \multirow{3}{*}{\makecell[c]{MVTec-\\3D AD}}
      & 100\% & 30\% & 100\% & 30\% & 0.675 & 0.655 & \underline{0.680} & 0.679 & 0.654 & \textcolor{deepred}{\textbf{0.707}} \\
      & & 30\% & 100\% & 30\% & 100\% & 0.662 & 0.642 & 0.664 & \underline{0.671} & 0.652 & \textcolor{deepred}{\textbf{0.703}} \\
      & & 65\% & 65\% & 65\% & 65\% & 0.668 & 0.648 & 0.672 & \underline{0.677} & 0.653 & \textcolor{deepred}{\textbf{0.706}} \\\cline{2-12}
      & \multirow{3}{*}{\makecell[c]{Eye-\\candies}}
      & 100\% & 30\% & 100\% & 30\% & 0.629 & 0.610 & 0.621 & \underline{0.634} & 0.615 & \textcolor{deepred}{\textbf{0.662}} \\
      & & 30\% & 100\% & 30\% & 100\% & 0.623 & 0.604 & 0.616 & \underline{0.624} & 0.607 & \textcolor{deepred}{\textbf{0.654}} \\
      & & 65\% & 65\% & 65\% & 65\% & 0.624 & 0.605 & 0.617 & \underline{0.626} & 0.610 & \textcolor{deepred}{\textbf{0.658}} \\
      
      \toprule[1.5pt]
    \end{tabular}
    }
    \label{tab:miiad_bench_with_incomplete_I-AUROC}
  \end{center}
  % \vspace{-15pt}
\end{table*}

\begin{table*}[t]
  \begin{center}
  \captionsetup{font={small,stretch=1.25}, labelfont={bf}}
  \caption{P-AUROC score in performance comparison of missing-modality based methods on \textbf{\textit{MIIAD Bench}}.}
  \vspace{-10pt}
   \renewcommand{\arraystretch}{1.2}
   \resizebox{1\textwidth}{!}{
    \begin{tabular}{c|c||c c||c c||c c c c c c}
     \toprule[1.5pt]
     
      \multirow{2}{*}[-0.5ex]{\makecell[c]{\textbf{Missing}\\\textbf{Rate}}} &
      \multirow{2}{*}[-0.5ex]{\makecell[c]{\textbf{\textit{MIIAD Bench}}\\\textbf{Split}}}  & 
      \multicolumn{2}{c||}{\textbf{Data in \# Train}} & 
      \multicolumn{2}{c||}{\textbf{Data in \# Test}} & 
      \multicolumn{6}{c}{\textbf{P-AUROC}} \\
      \cline{3-12}
      & & \textbf{3D PC} & \textbf{2D RGB} & \textbf{3D PC} & \textbf{2D RGB} & 
      \makecell[c]{\textbf{M3DM}\\\textbf{(Baseline)}}~\cite{wang2023multimodal} & \textbf{+GMC}~\cite{poklukar2022gmc} & \textbf{+VLMo}~\cite{bao2022vlmo} & \textbf{+ShaSpec}~\cite{wang2023multi} & \textbf{+CMDIAD}~\cite{sui2024incompletemultimodalindustrialanomaly} & \makecell[c]{\textbf{+RADAR}\\\textbf{(Ours)}}\\\hline\hline

      \multirow{2}{*}{0\%}
      & \makecell[c]{MVTec-3D AD} & \multicolumn{2}{c||}{100\%} & \multicolumn{2}{c||}{100\%} & \textcolor{deepred}{\textbf{0.992}} & 0.977 & 0.991 & \textcolor{deepred}{\textbf{0.992}} & 0.991 & \textcolor{deepred}{\textbf{0.992}} \\\cline{2-12}
      & \makecell[c]{Eyecandies} & \multicolumn{2}{c||}{100\%} & \multicolumn{2}{c||}{100\%} & 0.977 & 0.962 & 0.976 & \underline{0.978} & 0.977 & \textcolor{deepred}{\textbf{0.981}} \\
      \hline

      \multirow{6}{*}{30\%} 
      & \multirow{3}{*}{\makecell[c]{MVTec-\\3D AD}}
      & 100\% & 70\% & 100\% & 70\% & 0.945 & 0.929 & 0.943 & \underline{0.949} & 0.947 & \textcolor{deepred}{\textbf{0.955}} \\
      & & 70\% & 100\% & 70\% & 100\% & 0.938 & 0.922 & 0.936 & \underline{0.941} & 0.939 & \textcolor{deepred}{\textbf{0.948}} \\
      & & 85\% & 85\% & 85\% & 85\% & 0.942 & 0.926 & 0.940 & \underline{0.945} & 0.943 & \textcolor{deepred}{\textbf{0.953}} \\\cline{2-12}
      & \multirow{3}{*}{\makecell[c]{Eye-\\candies}}
      & 100\% & 70\% & 100\% & 70\% & 0.928 & 0.912 & 0.926 & \underline{0.933} & 0.930 & \textcolor{deepred}{\textbf{0.943}} \\
      & & 70\% & 100\% & 70\% & 100\% & 0.921 & 0.905 & 0.919 & \underline{0.926} & 0.923 & \textcolor{deepred}{\textbf{0.936}} \\
      & & 85\% & 85\% & 85\% & 85\% & 0.925 & 0.909 & 0.923 & \underline{0.930} & 0.927 & \textcolor{deepred}{\textbf{0.941}} \\
      \hline

      \multirow{6}{*}{50\%} 
      & \multirow{3}{*}{\makecell[c]{MVTec-\\3D AD}}
      & 100\% & 50\% & 100\% & 50\% & 0.865 & 0.849 & 0.863 & \underline{0.872} & 0.870 & \textcolor{deepred}{\textbf{0.888}} \\
      & & 50\% & 100\% & 50\% & 100\% & 0.857 & 0.841 & 0.855 & \underline{0.864} & 0.862 & \textcolor{deepred}{\textbf{0.880}} \\
      & & 75\% & 75\% & 75\% & 75\% & 0.861 & 0.845 & 0.859 & \underline{0.868} & 0.866 & \textcolor{deepred}{\textbf{0.885}} \\\cline{2-12}
      & \multirow{3}{*}{\makecell[c]{Eye-\\candies}}
      & 100\% & 50\% & 100\% & 50\% & 0.847 & 0.831 & 0.845 & \underline{0.853} & 0.851 & \textcolor{deepred}{\textbf{0.873}} \\
      & & 50\% & 100\% & 50\% & 100\% & 0.839 & 0.823 & 0.837 & \underline{0.845} & 0.843 & \textcolor{deepred}{\textbf{0.865}} \\
      & & 75\% & 75\% & 75\% & 75\% & 0.843 & 0.827 & 0.841 & \underline{0.849} & 0.847 & \textcolor{deepred}{\textbf{0.870}} \\
      \hline

      \multirow{6}{*}{70\%} 
      & \multirow{3}{*}{\makecell[c]{MVTec-\\3D AD}}
      & 100\% & 30\% & 100\% & 30\% & 0.755 & 0.740 & 0.753 & \underline{0.760} & 0.758 & \textcolor{deepred}{\textbf{0.785}} \\
      & & 30\% & 100\% & 30\% & 100\% & 0.747 & 0.732 & 0.745 & \underline{0.752} & 0.750 & \textcolor{deepred}{\textbf{0.777}} \\
      & & 65\% & 65\% & 65\% & 65\% & 0.751 & 0.736 & 0.749 & \underline{0.756} & 0.754 & \textcolor{deepred}{\textbf{0.782}} \\\cline{2-12}
      & \multirow{3}{*}{\makecell[c]{Eye-\\candies}}
      & 100\% & 30\% & 100\% & 30\% & 0.737 & 0.722 & 0.735 & \underline{0.742} & 0.740 & \textcolor{deepred}{\textbf{0.763}} \\
      & & 30\% & 100\% & 30\% & 100\% & 0.729 & 0.714 & 0.727 & \underline{0.734} & 0.732 & \textcolor{deepred}{\textbf{0.755}} \\
      & & 65\% & 65\% & 65\% & 65\% & 0.733 & 0.718 & 0.731 & \underline{0.738} & 0.736 & \textcolor{deepred}{\textbf{0.760}} \\
      
      \toprule[1.5pt]
    \end{tabular}
    }
    \label{tab:miiad_bench_with_incomplete_P-AUROC}
  \end{center}
  % \vspace{-15pt}
\end{table*}

\begin{table*}[t]
  \begin{center}
  \captionsetup{font={small,stretch=1.25}, labelfont={bf}}
  \caption{AUPRO score in performance comparison of missing-modality based methods on \textbf{\textit{MIIAD Bench}}.}
  \vspace{-10pt}
   \renewcommand{\arraystretch}{1.2}
   \resizebox{1\textwidth}{!}{
    \begin{tabular}{c|c||c c||c c||c c c c c c}
     \toprule[1.5pt]
     
      \multirow{2}{*}[-0.5ex]{\makecell[c]{\textbf{Missing}\\\textbf{Rate}}} &
      \multirow{2}{*}[-0.5ex]{\makecell[c]{\textbf{\textit{MIIAD Bench}}\\\textbf{Split}}}  & 
      \multicolumn{2}{c||}{\textbf{Data in \# Train}} & 
      \multicolumn{2}{c||}{\textbf{Data in \# Test}} & 
      \multicolumn{6}{c}{\textbf{AUPRO}} \\
      \cline{3-12}
      & & \textbf{3D PC} & \textbf{2D RGB} & \textbf{3D PC} & \textbf{2D RGB} & 
      \makecell[c]{\textbf{M3DM}\\\textbf{(Baseline)}}~\cite{wang2023multimodal} & \textbf{+GMC}~\cite{poklukar2022gmc} & \textbf{+VLMo}~\cite{bao2022vlmo} & \textbf{+ShaSpec}~\cite{wang2023multi} & \textbf{+CMDIAD}~\cite{sui2024incompletemultimodalindustrialanomaly} & \makecell[c]{\textbf{+RADAR}\\\textbf{(Ours)}}\\\hline\hline

      \multirow{2}{*}{0\%}
      & \makecell[c]{MVTec-3D AD} & \multicolumn{2}{c||}{100\%} & \multicolumn{2}{c||}{100\%} & 0.964 & 0.952 & \underline{0.965} & 0.958 & 0.960 & \textcolor{deepred}{\textbf{0.967}} \\\cline{2-12}
      & \makecell[c]{Eyecandies} & \multicolumn{2}{c||}{100\%} & \multicolumn{2}{c||}{100\%} & 0.882 & 0.871 & 0.883 & 0.882 & \underline{0.884} & \textcolor{deepred}{\textbf{0.885}} \\
      \hline

      \multirow{6}{*}{30\%} 
      & \multirow{3}{*}{\makecell[c]{MVTec-\\3D AD}}
      & 100\% & 70\% & 100\% & 70\% & 0.911 & 0.895 & 0.910 & \underline{0.915} & 0.911 & \textcolor{deepred}{\textbf{0.925}} \\
      & & 70\% & 100\% & 70\% & 100\% & 0.932 & 0.915 & 0.934 & \underline{0.938} & 0.916 & \textcolor{deepred}{\textbf{0.947}} \\
      & & 85\% & 85\% & 85\% & 85\% & 0.917 & 0.902 & \underline{0.918} & 0.917 & 0.913 & \textcolor{deepred}{\textbf{0.929}} \\\cline{2-12}
      & \multirow{3}{*}{\makecell[c]{Eye-\\candies}}
      & 100\% & 70\% & 100\% & 70\% & 0.828 & 0.813 & 0.827 & \underline{0.834} & 0.831 & \textcolor{deepred}{\textbf{0.847}} \\
      & & 70\% & 100\% & 70\% & 100\% & 0.832 & 0.817 & 0.835 & \underline{0.841} & 0.838 & \textcolor{deepred}{\textbf{0.851}} \\
      & & 85\% & 85\% & 85\% & 85\% & 0.830 & 0.815 & 0.833 & \underline{0.840} & 0.835 & \textcolor{deepred}{\textbf{0.854}} \\
      \hline

      \multirow{6}{*}{50\%} 
      & \multirow{3}{*}{\makecell[c]{MVTec-\\3D AD}}
      & 100\% & 50\% & 100\% & 50\% & 0.835 & 0.817 & 0.833 & \underline{0.843} & 0.840 & \textcolor{deepred}{\textbf{0.857}} \\
      & & 50\% & 100\% & 50\% & 100\% & 0.849 & 0.831 & 0.847 & \underline{0.856} & 0.854 & \textcolor{deepred}{\textbf{0.874}} \\
      & & 75\% & 75\% & 75\% & 75\% & 0.833 & 0.815 & 0.830 & \underline{0.839} & 0.837 & \textcolor{deepred}{\textbf{0.862}} \\\cline{2-12}
      & \multirow{3}{*}{\makecell[c]{Eye-\\candies}}
      & 100\% & 50\% & 100\% & 50\% & 0.749 & 0.734 & 0.746 & \underline{0.755} & 0.753 & \textcolor{deepred}{\textbf{0.772}} \\
      & & 50\% & 100\% & 50\% & 100\% & 0.741 & 0.726 & 0.738 & \underline{0.748} & 0.746 & \textcolor{deepred}{\textbf{0.769}} \\
      & & 75\% & 75\% & 75\% & 75\% & 0.742 & 0.727 & 0.739 & \underline{0.749} & 0.747 & \textcolor{deepred}{\textbf{0.771}} \\
      \hline

      \multirow{6}{*}{70\%} 
      & \multirow{3}{*}{\makecell[c]{MVTec-\\3D AD}}
      & 100\% & 30\% & 100\% & 30\% & 0.715 & 0.703 & 0.710 & \underline{0.717} & 0.703 & \textcolor{deepred}{\textbf{0.755}} \\
      & & 30\% & 100\% & 30\% & 100\% & 0.709 & 0.708 & 0.713 & \underline{0.718} & 0.712 & \textcolor{deepred}{\textbf{0.748}} \\
      & & 65\% & 65\% & 65\% & 65\% & 0.711 & 0.701 & 0.711 & \underline{0.717} & 0.707 & \textcolor{deepred}{\textbf{0.752}} \\\cline{2-12}
      & \multirow{3}{*}{\makecell[c]{Eye-\\candies}}
      & 100\% & 30\% & 100\% & 30\% & 0.613 & 0.597 & 0.611 & \underline{0.614} & 0.606 & \textcolor{deepred}{\textbf{0.647}} \\
      & & 30\% & 100\% & 30\% & 100\% & 0.608 & 0.592 & 0.605 & \underline{0.611} & 0.598 & \textcolor{deepred}{\textbf{0.649}} \\
      & & 65\% & 65\% & 65\% & 65\% & 0.610 & 0.598 & 0.608 & \underline{0.613} & 0.601 & \textcolor{deepred}{\textbf{0.649}} \\
      
      \toprule[1.5pt]
    \end{tabular}
    }
    \label{tab:miiad_bench_with_incomplete_AUPRO}
  \end{center}
  % \vspace{-15pt}
\end{table*}

\begin{table*}[t]
  \begin{center}
  \captionsetup{font={small,stretch=1.25}, labelfont={bf}}
  \caption{Ablation study of different modules under varying missing rates on \textit{MIIAD Bench} - MVTec-3D AD. Here I-A. represents I-AUROC, AUP. represents AUPRO.}
  \vspace{-10pt}
   \renewcommand{\arraystretch}{1.2}
   \resizebox{\textwidth}{!}{
     \begin{tabular}{l|c c c||c c|c c|c c||c c|c c|c c}
      \toprule[1.5pt]
      
      \multirow{2}{*}{\textbf{Methods}} & \multicolumn{3}{c||}{\textbf{Setting}} & 
      \multicolumn{2}{c}{\textbf{30\% PC}} & 
      \multicolumn{2}{c}{\textbf{50\% PC}} & 
      \multicolumn{2}{c||}{\textbf{70\% PC}} & 
      \multicolumn{2}{c}{\textbf{30\% RGB}} & 
      \multicolumn{2}{c}{\textbf{50\% RGB}} & 
      \multicolumn{2}{c}{\textbf{70\% RGB}}\\
      \cline{2-16}
      & MII & ALM & DPHM & I-A. & AUP. & I-A. & AUP. & I-A. & AUP. & I-A. & AUP. & I-A. & AUP. & I-A. & AUP.\\
      \hline \hline

      Baseline & & & & 0.883 & 0.932 & 0.789 & 0.849 & 0.662 & 0.709 & 0.892 & 0.911 & 0.796 & 0.835 & 0.675 & 0.715\\
      + MII & \faCheckCircle & & & 0.892 & 0.940 & 0.802 & 0.860 & 0.680 & 0.725 & 0.899 & 0.918 & 0.808 & 0.845 & 0.690 & 0.728\\
      + ALM & & \faCheckCircle & & 0.885 & 0.934 & 0.795 & 0.853 & 0.668 & 0.714 & 0.893 & 0.914 & 0.801 & 0.839 & 0.683 & 0.721\\
      + DPHM & & & \faCheckCircle & 0.889 & 0.937 & 0.798 & 0.856 & 0.675 & 0.719 & 0.896 & 0.917 & 0.805 & 0.842 & 0.687 & 0.725\\
      + MII \& ALM & \faCheckCircle & \faCheckCircle & & 0.898 & 0.943 & 0.810 & 0.865 & 0.688 & 0.732 & 0.905 & 0.923 & 0.816 & 0.851 & 0.698 & 0.736\\
      + MII \& DPHM & \faCheckCircle & & \faCheckCircle & \underline{0.902} & \underline{0.945} & \underline{0.816} & \underline{0.868} & \underline{0.693} & \underline{0.738} & \underline{0.907} & \textcolor{deepred}{\textbf{0.925}} & \underline{0.823} & \underline{0.858} & \underline{0.705} & \underline{0.743}\\
      + ALM \& DPHM & & \faCheckCircle & \faCheckCircle & 0.894 & 0.940 & 0.804 & 0.862 & 0.682 & 0.727 & 0.901 & 0.921 & 0.810 & 0.848 & 0.693 & 0.733\\
      \textbf{RADAR (Ours)} & \faCheckCircle & \faCheckCircle & \faCheckCircle & \textcolor{deepred}{\textbf{0.903}} & \textcolor{deepred}{\textbf{0.947}} & \textcolor{deepred}{\textbf{0.821}} & \textcolor{deepred}{\textbf{0.874}} & \textcolor{deepred}{\textbf{0.703}} & \textcolor{deepred}{\textbf{0.748}} & \textcolor{deepred}{\textbf{0.908}} & \textcolor{deepred}{\textbf{0.925}} & \textcolor{deepred}{\textbf{0.830}} & \textcolor{deepred}{\textbf{0.865}} & \textcolor{deepred}{\textbf{0.715}} & \textcolor{deepred}{\textbf{0.755}}\\
      \toprule[1.5pt]
     \end{tabular}
     }
     \label{tab:ablation_study_more}
  \end{center}
  \vspace{-5pt}
\end{table*}

\begin{table*}[t]
  \begin{center}
  \captionsetup{font={small,stretch=1.25}, labelfont={bf}}
  \caption{Ablation study of dual-branch architecture on \textit{MIIAD Bench} - MVTec-3D AD. Performance reported in I-AUROC. Here ``others'' represents other modules.}
  \vspace{-10pt}
   \renewcommand{\arraystretch}{1.2}
   \resizebox{0.57\textwidth}{!}{
     \begin{tabular}{l|c|c|c|c}
      \toprule[1.5pt]
      \textbf{Configurations} & \textbf{30\% PC} & \textbf{70\% PC} & \textbf{30\% RGB} & \textbf{70\% RGB} \\
      \hline \hline
      Baseline & 0.883 & 0.662 & 0.892 & 0.675 \\
      \rowcolor{gray!10} w Branch 1 (only) & 0.886 & 0.670 & 0.894 & 0.684 \\
      w Branch 2 (only) & 0.888 & 0.671 & 0.894 & 0.683 \\
      \rowcolor{gray!10} w Branch 1 + others & 0.899 & 0.693 & 0.905 & 0.706 \\
      w Branch 2 + others & \underline{0.901} & \underline{0.698} & \underline{0.907} & \underline{0.710} \\
      \rowcolor{gray!10} \textbf{All modules (Ours)} & \textcolor{deepred}{\textbf{0.903}} & \textcolor{deepred}{\textbf{0.703}} & \textcolor{deepred}{\textbf{0.908}} & \textcolor{deepred}{\textbf{0.715}} \\
      \toprule[1.5pt]
     \end{tabular}
     }
     \label{tab:branch_ablation}
  \end{center}
  \vspace{-10pt}
\end{table*}

\begin{figure*}[!h]
\includegraphics[width=1\textwidth]{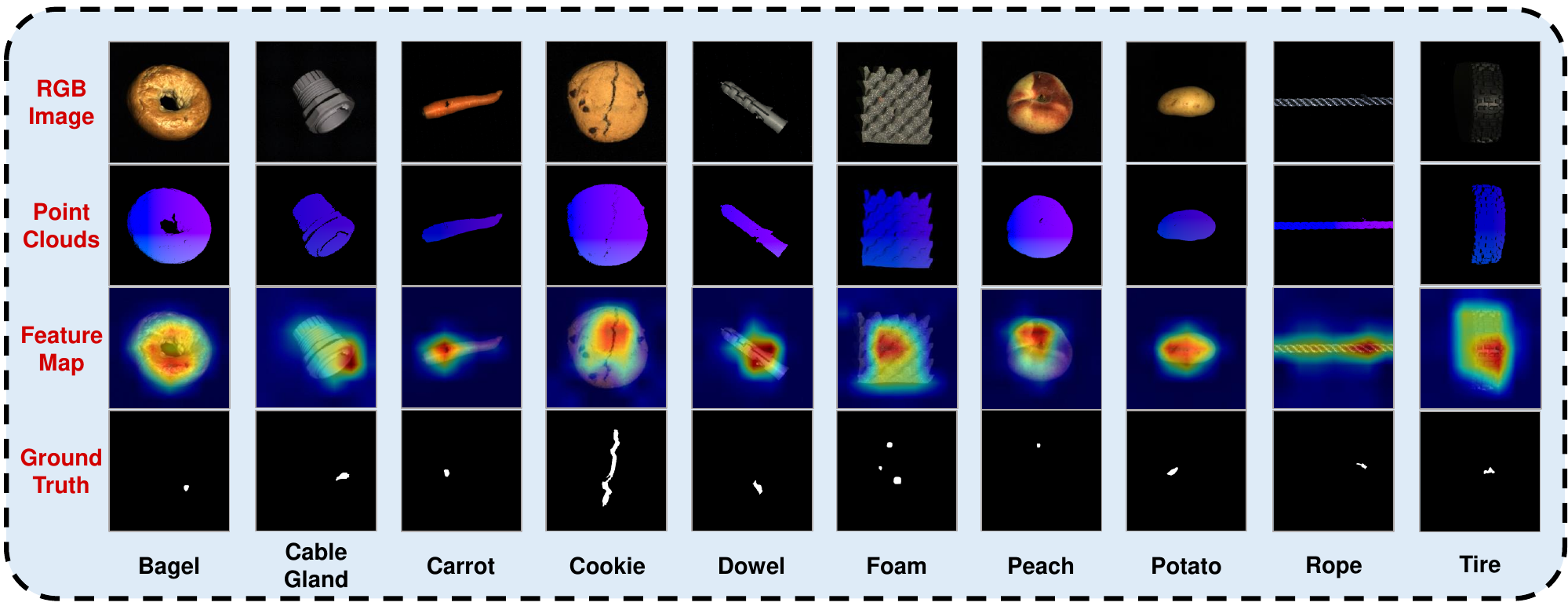}
\centering\caption{Additional visualization results of industrial anomaly detection.} 
\label{fig:additional_visualization_results}
\end{figure*}

\begin{table*}[t]
  \begin{center}
  \captionsetup{font={small,stretch=1.25}, labelfont={bf}}
  \caption{Ablation study of $\lambda$ parameters on \textit{MIIAD Bench} - MVTec-3D AD. Here I-A. represents I-AUROC, AUP. represents AUPRO.}
  \vspace{-10pt}
   \renewcommand{\arraystretch}{1.2}
   \resizebox{0.7\textwidth}{!}{
     \begin{tabular}{l|c c||c c|c c|c c}
      \toprule[1.5pt]
      
      \multirow{2}{*}{\textbf{Configurations}} & 
      \multicolumn{2}{c||}{\textbf{Parameters}} & 
      \multicolumn{2}{c}{\textbf{30\% PC}} & 
      \multicolumn{2}{c}{\textbf{50\% PC}} & 
      \multicolumn{2}{c}{\textbf{70\% PC}}\\
      \cline{2-9}
      & $\lambda_{rgb}$ & $\lambda_{pc}$ & I-A. & AUP. & I-A. & AUP. & I-A. & AUP.\\
      \hline \hline

      Baseline & 0.0 & 0.0 & 0.898 & 0.943 & 0.810 & 0.865 & 0.688 & 0.732 \\
      \rowcolor{gray!10} $\lambda_{pc}$ Only & 0.0 & 0.5 & 0.894 & 0.939 & 0.804 & 0.858 & 0.679 & 0.724 \\
      $\lambda_{rgb}$ Only & 1.0 & 0.0 & 0.897 & 0.942 & 0.805 & 0.859 & 0.683 & 0.726 \\
      \rowcolor{gray!10} Balanced & 0.7 & 0.3 & 0.899 & 0.945 & 0.812 & 0.863 & 0.691 & 0.734 \\
      Mixed & 1.0 & 0.3 & \underline{0.901} & \textcolor{deepred}{\textbf{0.947}} & \underline{0.816} & \underline{0.867} & \underline{0.697} & \underline{0.739} \\
      \rowcolor{gray!10} \textbf{Optimal (Ours)} & \textbf{1.0} & \textbf{0.5} & \textcolor{deepred}{\textbf{0.903}} & \textcolor{deepred}{\textbf{0.947}} & \textcolor{deepred}{\textbf{0.821}} & \textcolor{deepred}{\textbf{0.874}} & \textcolor{deepred}{\textbf{0.703}} & \textcolor{deepred}{\textbf{0.748}} \\
      Overweight PC & 0.5 & 1.0 & 0.895 & 0.940 & 0.807 & 0.858 & 0.685 & 0.729 \\
      \toprule[1.5pt]
     \end{tabular}
     }
     \label{tab:lambda_ablation}
  \end{center}
  \vspace{-10pt}
\end{table*}

\section{Comprehensive Ablation Study}
\label{sec:ablation}

Tab.~\ref{tab:ablation_study_more} presents more comprehensive ablation results, showing the relationship between model components and I-AUROC/AUPRO metrics under various modality missing scenarios: $30\%$ PC, $50\%$ PC, $70\%$ PC, $30\%$ RGB, $50\%$ RGB, and $70\%$ RGB missing rates. 

The experimental results reconfirm that all three components in our framework effectively enhance baseline performance, with Modality-incomplete Instruction (MII) demonstrating the most significant impact, followed by Double-Pseudo Hybrid Module (DPHM), and finally Adaptive Learning Module (ALM).

\section{Ablation Study of Double-Branch}
\label{sec:ablation_double-branch}

The ablation study on dual-branch architecture (Tab.~\ref{tab:branch_ablation}) reveals several key insights. When implemented individually, both branches demonstrate measurable performance gains over the baseline, with Branch 2 showing slightly stronger results in point cloud missing scenarios ($0.888$ vs $0.886$ at 30\% PC) while Branch 1 performs marginally better in RGB missing conditions ($0.684$ vs $0.683$ at $70\%$ RGB). This suggests specialized feature extraction capabilities: Branch 2 appears more sensitive to geometric information while Branch 1 better captures texture patterns.

Significantly greater improvements emerge when integrating each branch with complementary modules. The addition of Branch 1 boosts performance by $1.6-3.1\%$ across test conditions, while Branch 2 delivers even stronger gains of $1.8-3.5\%$. However, the full integration of both branches with all modules achieves optimal results, particularly excelling under challenging $70\%$ missing rates with $4.1\%$ improvement for point cloud and $4.0\%$ for RGB modalities. This demonstrates that while each branch contributes valuable representations, their synergistic combination is essential for maximal performance, confirming the architectural design's effectiveness in handling multimodal missing data scenarios.

\section{Ablation Study of Hyper-Parameters}
\label{sec:hyper_parameters}

To systematically analyze the impact of $\lambda_{rgb}$ and $\lambda_{pc}$ in the reconstruction loss of Stage2 Branch 1, we conduct controlled experiments with seven parameter configurations under varying point cloud (PC) missing rates ($30\%$, $50\%$, $70\%$). The experimental setup includes: (1) \textit{Baseline} without weight adjustment ($\lambda_{rgb}=0$, $\lambda_{pc}=0$), (2) 3D-only weighting ($\lambda_{rgb}=0$, $\lambda_{pc}=0.5$), (3) RGB-only emphasis ($\lambda_{rgb}=1.0$, $\lambda_{pc}=0$), (4) Balanced weights ($\lambda_{rgb}=0.7$, $\lambda_{pc}=0.3$), (5) Asymmetric combination ($\lambda_{rgb}=1.0$, $\lambda_{pc}=0.3$), (6) Our optimal configuration ($\lambda_{rgb}=1.0$, $\lambda_{pc}=0.5$), and (7) 3D-overweighted scenario ($\lambda_{rgb}=0.5$, $\lambda_{pc}=1.0$). 

Experimental results are shown in Tab.~\ref{tab:lambda_ablation}, which reveal that the optimal configuration ($\lambda_{rgb}=1.0$ + $\lambda_{pc}=0.5$) achieves consistent performance gains across all test conditions, with I-AUROC improvements of $+1.1\%$ ($50\%$ PC missing rate) and $+1.5\%$ ($70\%$ PC missing rate) compared to baseline. 

Notably, the $\lambda_{rgb}$-only configuration shows better robustness than $\lambda_{pc}$-only at $30\%$ missing rate ($+0.3\%$ I-AUROC, $0.897$ vs $0.894$), indicating RGB modality's stronger discriminative power in partial observation scenarios. However, both single-modality configurations underperform the baseline, demonstrating the necessity of multimodal integration.

The synergistic effect of combined weights yields maximum performance, where 3D weighting provides complementary geometric constraints without overriding RGB's texture sensitivity. The performance degradation in 3D-overweighted configuration ($-0.3\%$ I-AUROC at $70\%$ missing rate compared to baseline) further validates the importance of maintaining modality balance.

\section{Additional Visualization Results}
\label{sec:more_visualization_results}
As depicted in Fig. \ref{fig:additional_visualization_results}, we present additional industrial anomaly results for all categories within the \texttt{MIIAD Dataset}. As can be seen from the figure, for each category of the dataset, our model can identify abnormal industrial entities while accurately locating the position of anomalies.

% \clearpage

% \balance
% \bibliographystyle{ACM-Reference-Format}
% \bibliography{reference}

% \appendix









\end{document}